\documentclass[final]{template1}
\usepackage[most]{tcolorbox}
\usepackage{times,booktabs,microtype,mathtools}
\usepackage{enumitem,threeparttable}
\setlist[itemize]{leftmargin=*,itemsep=3pt,topsep=4pt}
\definecolor{checkgreen}{RGB}{34,139,34}
\newcommand{\greencheck}{\textcolor{checkgreen}{\ensuremath{\checkmark}}}
\DontPrintSemicolon
\SetCommentSty{textrm}
\usepackage[capitalize,noabbrev]{cleveref}
\newtheorem{indlemma}{Lemma}
\newtheorem{indproposition}{Proposition}
\newtheorem{indcorollary}{Corollary}
\newtheorem{indassumption}{Assumption}
\theorembodyfont{\normalfont}
\newtheorem{indremark}{Remark}
\theorembodyfont{\normalfont\itshape}

\crefname{theorem}{Theorem}{Theorems}
\Crefname{theorem}{Theorem}{Theorems}
\crefname{indlemma}{Lemma}{Lemmas}
\Crefname{indlemma}{Lemma}{Lemmas}
\crefname{indproposition}{Proposition}{Propositions}
\Crefname{indproposition}{Proposition}{Propositions}
\crefname{indcorollary}{Corollary}{Corollaries}
\Crefname{indcorollary}{Corollary}{Corollaries}
\crefname{indassumption}{Assumption}{Assumptions}
\Crefname{indassumption}{Assumption}{Assumptions}
\crefname{indremark}{Remark}{Remarks}
\Crefname{indremark}{Remark}{Remarks}
\newcommand{\R}{\mathbb R}
\newcommand{\eps}{\varepsilon}
\newcommand{\ip}[2]{\langle #1,#2\rangle}
\newcommand{\norm}[1]{\lVert #1\rVert}
\newcommand{\F}{\mathrm F}
\newcommand{\op}{\mathrm{op}}
\newcommand{\diag}{\operatorname{diag}}
\newcommand{\Id}{\mathrm I}
\newcommand{\polar}{\operatorname{polar}}
\newcommand{\argmin}{\operatorname*{arg\,min}}
\newcommand{\argmax}{\operatorname*{arg\,max}}
\newcommand{\cL}{\mathcal L}
\title[When Muon Loses Its Spectral Implicit Bias]{From Spectral to Frobenius:\\ When Muon Loses Its Spectral Implicit Bias}
\usepackage{times}
\atauthor{%
 \Name{Yongtao Wu} 
  \Email{yongtaowu@ruc.edu.cn}
 \\
 \addr Renmin University of China
 \AND
 \Name{Mahdi Mansouri} \Email{uni.mahdi.mansouri@gmail.com}
 \\
 \addr The University of Hong Kong
 \AND
 \Name{Zhenyu Zhu} 
  \Email{zzhenyu@hku.hk}
 \\
 \addr The University of Hong Kong%
}

\begin{document}

\maketitle

\begingroup
\renewcommand{\thefootnote}{*}
\footnotetext[1]{Correspondence to:
zzhenyu@hku.hk.}
\endgroup

\begin{abstract}
Prior work on implicit bias shows that gradient descent selects the Frobenius max-margin direction on linearly separable problems, whereas \citet{fan2025spectral} established spectral max-margin behavior for spectral-norm steepest descent methods, including idealized Muon with exact orthogonalization. In practice, however, Muon approximates orthogonalization using finite Newton–Schulz iterations and a positive stabilization constant. We prove that these implementation choices cause Muon to select the Frobenius rather than the spectral max-margin direction in linear classification with separable data. We also establish this behavior for both exponential moving average and Nesterov-style momentum. The key technical ingredient is a summable perturbation argument showing that the resulting parameter trajectories remain at bounded distance from those of gradient descent. We then disentangle the roles of Newton–Schulz iteration count, the stabilization constant, and the Newton–Schulz polynomial, revealing distinct asymptotic behaviors, including directions that achieve neither Frobenius nor spectral max-margin. Finally, we show that Muon can recover its spectral max-margin bias with Newton–Schulz iterations growing logarithmically over training.
\end{abstract}

\section{Introduction}
Optimization algorithms can favor different solutions even when minimizing the same training objective, a phenomenon known as implicit bias \citep{neyshabur2015search,gunasekar2017matrix}. A classical example arises in linear classification with separable data, where logistic and exponential losses can approach zero only as the parameter norm diverges. In this setting, the limiting direction of the parameters becomes the object of interest. Gradient descent converges in direction to the Euclidean max-margin classifier \citep{soudry2018implicit,nacson2019convergence,ji2020regularization,ji2021primaldual}, whereas steepest descent under different norms can select different max-margin solutions \citep{gunasekar2018geometry,wang2023faster}. 

Muon, an optimizer that has attracted considerable attention for large language model training \citep{jordan2024muon,liu2025scalable}, orthogonalizes matrix-valued updates rather than following conventional gradient directions. Ignoring momentum for simplicity, its idealized update maps a gradient matrix $G=U\Sigma V^\top$ to its polar factor $UV^\top$. This update corresponds to steepest descent under the spectral norm, in contrast to the Frobenius-norm geometry of standard gradient descent. Recent work has studied the implicit bias of Muon with exact orthogonalization, establishing spectral max-margin behavior in linear multiclass classification \citep{fan2025spectral,li2026minibatch} and margin stationarity in homogeneous neural networks \citep{gronich2026homogeneous}.

However, these analyses rely on exact orthogonalization, which can be computationally expensive for large matrices. Practical Muon instead approximates the polar factor using a fixed number of Newton--Schulz iterations. Specifically, the original Muon optimizer proposed by \citet{jordan2024muon}, when considered without momentum, normalizes the gradient as $G/(\norm{G}_{\F}+\eps)$, with a fixed stabilization constant $\eps>0$, and then applies five Newton-Schulz iterations using a quintic polynomial. Whether these practical modifications preserve the spectral max-margin bias of idealized Muon remains unclear. This raises the central question of this work:

% \begin{center}\itshape
% Does practical Muon retain the spectral max-margin bias of its idealized counterpart?
% \end{center}
\begin{tcolorbox}[
    colback=black!5,
    colframe=black!60,
    boxrule=0.8pt,
    arc=5pt,
    left=8pt,
    right=8pt,
    top=6pt,
    bottom=6pt,
    before skip=10pt,
    after skip=10pt
]
\centering\itshape
Does practical Muon retain the spectral max-margin bias of its idealized counterpart?
\end{tcolorbox}

We show that the answer is negative: practical Muon with a fixed number of Newton-Schulz iterations and positive stabilization selects the Frobenius rather than the spectral max-margin direction on linear classification problems. The key technical idea is to express the practical Muon update as a gradient descent step plus a summable perturbation, thereby establishing that their parameter trajectories remain at bounded distance. We further investigate how the Newton–Schulz iteration count, stabilization constant, and polynomial choice affect the implicit bias and establish conditions under which spectral max-margin behavior can be recovered. Our contributions are as follows.
\begin{itemize}
\item We establish that Muon with a fixed number of Newton-Schulz iterations and positive stabilization leads to Frobenius max-margin selection on linear classification problems. Our results cover a general class of Newton–Schulz polynomials, including the quintic polynomial used in practical Muon, and extend to both exponential moving average and Nesterov-style momentum (\cref{sec:main}).
% \item \textbf{Frobenius max-margin bias of practical Muon.} We establish that practical Muon with a fixed number of Newton-Schulz iterations and positive stabilization lead to Frobenius max-margin selection on linear classification problems. Our results cover a general class of Newton–Schulz polynomials, including the quintic polynomial used in practical Muon, and extend to both exponential moving average and Nesterov-style momentum (\cref{sec:main}).

\item We examine how the Newton–Schulz iteration count, stabilization constant, and polynomial choice affect implicit bias. We show that removing stabilization alone does not necessarily restore spectral max-margin behavior: a fixed number of Newton-Schulz iterations can select directions that achieve neither Frobenius nor spectral max-margin. We further characterize the limiting behavior as the Newton–Schulz iteration count tends to infinity, revealing distinct asymptotic behaviors for the cubic and quintic polynomials (\cref{sec:limits}).

% \item \textbf{Disentangling implementation choices.} We examine how the Newton–Schulz iteration count, stabilization constant, and polynomial choice affect implicit bias. We show that removing stabilization alone does not necessarily restore spectral max-margin behavior: a fixed number of Newton-Schulz iterations can select directions that achieve neither Frobenius nor spectral max-margin. We further characterize the limiting behavior as the Newton–Schulz iteration count tends to infinity, revealing distinct asymptotic behaviors for the cubic and quintic polynomials (\cref{sec:limits}).

\item We show that spectral max-margin behavior can be recovered with logarithmically growing Newton-Schulz iteration counts during training. For the cubic polynomial, this holds with a fixed positive stabilization constant, while for the quintic polynomial, we establish a recovery guarantee using gradient-dependent stabilization (\cref{sec:recovery}).

% \item \textbf{Recovery of spectral max-margin bias.} We show that spectral max-margin behavior can be recovered with logarithmically growing Newton-Schulz iteration counts during training. For the cubic polynomial, this holds with a fixed positive stabilization constant, while for the quintic polynomial, we establish a recovery guarantee using gradient-dependent stabilization (\cref{sec:recovery}).
\end{itemize}

\section{Preliminary}\label{sec:setup}
We introduce the matrix notation and separable classification problem, then specify the exact and practical Muon updates used throughout the paper.
\subsection{Notation}
For matrices of the same shape, let
$\ip A B=\operatorname{tr}(A^\top B)$,
where $\operatorname{tr}$ denotes the matrix trace. We use $\norm{\cdot}_{\F}$, $\norm{\cdot}_{\op}$, and $\norm{\cdot}_*$ to denote the Frobenius, spectral (i.e., Euclidean-induced operator), and nuclear norms, respectively. A compact singular-value decomposition (SVD) $G=U\Sigma V^\top$ contains the positive singular values, and $\polar(G)=UV^\top$, with $\polar(0)=0$.
For an odd polynomial $p$, write $p^{\circ K}$ for its $K$-fold composition, with $p^{\circ0}(s)=s$. For a compact SVD, $G=U\Sigma V^\top$, define $\mathcal P_K(G)=Up^{\circ K}(\Sigma)V^\top$, where $p^{\circ K}$ is applied entrywise to the diagonal entries of $\Sigma$. A detailed summary of the core notation can be found in Table~\ref{tab:notation}.

\subsection{Problem setting and implicit bias}\label{sec:problem}
Consider a binary classification dataset
$\{(X_i,y_i)\}_{i=1}^n$ with $n\in\mathbb{N}_{>0}$,
where $X_i\in\R^{d_1\times d_2}$ and
$y_i\in\{-1,+1\}$ denote the features and labels,
respectively. We consider linear predictors parameterized by
$W\in\R^{d_1\times d_2}$, with prediction scores
$\ip{X_i}{W}$.
Defining the signed features $A_i=y_iX_i$, the classification
margins are $\ip{A_i}{W}$.
The empirical loss is
\begin{equation}
    \cL(W)=\frac{1}{n}\sum_i\ell(\ip{A_i}{W}),
\label{equ:empiricalloss}
\end{equation}
with either logistic loss $\ell(z)=\log(1+e^{-z})$
or exponential loss $\ell(z)=e^{-z}$.
These are the standard exponential-tail losses used in implicit-bias analysis
\citep{soudry2018implicit,gunasekar2018geometry}. We assume that the data are linearly separable, as formalized below.

\begin{assumption}[Separable data]\label{ass:separable}
There exist $U\in\R^{d_1\times d_2}$ with
$\norm U_{\F}=1$ and $\gamma>0$ such that
$\ip{A_i}{U}\geq\gamma$ for every $i$.
\end{assumption}
\begin{remark}
This is the standard linear separability assumption used by
\citet{soudry2018implicit,gunasekar2018geometry} and,
in its multiclass form, by \citet{fan2025spectral}.
\end{remark}
Under Assumption~\ref{ass:separable}, the Frobenius
hard-margin separator is uniquely defined by
\begin{equation}\label{eq:hardmargin}
\widehat W_{\F}
=\argmin_W \norm W_{\F}
\quad\text{subject to }
\ip{A_i}{W}\geq1\quad(i\in[n]).
\end{equation}
Implicit bias describes the limiting direction selected
by an optimization algorithm as training progresses.
For gradient descent with a sufficiently small constant
stepsize, the iterates satisfy $
\lim_{t\to\infty}
\frac{W_t}{\norm{W_t}_{\F}}
=
\frac{\widehat W_{\F}}{\norm{\widehat W_{\F}}_{\F}},$ where $W_t$ denotes the parameter iterate at iteration $t$
\citep{soudry2018implicit}. In contrast, \citet{fan2025spectral} establish spectral
max-margin behavior for spectral-norm steepest descent,
including Muon with exact orthogonalization.
Analogous to \cref{eq:hardmargin}, a spectral hard-margin separator satisfies
\begin{equation}\label{eq:spectral-hardmargin}
\widehat W_{\op}
\in\argmin_W \norm W_{\op}
\quad\text{subject to }
\ip{A_i}{W}\geq1\quad(i\in[n]).
\end{equation}
The Frobenius max-margin $\gamma_{\F}$ and spectral max-margin $\gamma_{\op}$ are defined by
\begin{equation}\label{eq:spectral-margin}
\gamma_{\nu}
=\max_{\norm W_{\nu}\leq1}\min_{i\in[n]}\ip{A_i}{W}
=\frac{1}{\norm{\widehat W_{\nu}}_{\nu}},
\qquad \nu\in\{\F,\op\}.
\end{equation}
Under~\cref{ass:separable}, the max-margin
and hard-margin formulations yield the same optimal
directions. Specifically, normalizing any hard-margin
solution $\widehat W_\nu$ by its $\nu$-norm gives a
max-margin solution.
\subsection{Idealized and Practical Muon}\label{sec:muon}
Spectral-norm steepest descent uses the polar factor to maximize alignment with the gradient over the spectral-norm unit ball,
\begin{equation}\label{eq:spectral-oracle}
\polar(G)\in\argmax_{\norm D_{\op}\leq1}\ip{G}{D}.
\end{equation}
Idealized Muon applies this operation to an exponential moving average of gradients, as specified in \cref{alg:ideal}.

Practical Muon\footnote{Throughout this paper, ``practical Muon'' refers to the finite Newton--Schulz update, stabilization, and momentum, excluding additional components such as weight decay and update scaling.} replaces the SVD by finite matrix-polynomial iterations. The implementation of \citet{jordan2024muon} uses an EMA buffer and a Nesterov-style momentum input.
\begin{equation}\label{eq:practical-momentum}
M_t=\beta M_{t-1}+(1-\beta)G_t,\qquad
N_t=(1-\kappa)G_t+\kappa M_t,\qquad M_{-1}=0,
\end{equation}
where $G_t=\nabla\cL(W_t)$, $0\leq\beta<1$, and $0\leq\kappa\leq1$. The input $N_t$ is normalized by $\norm{N_t}_{\F}+\eps$ before applying $K$ polynomial iterations, as detailed in \cref{alg:practical}. The default practical Muon settings are $K=5$, $\beta=\kappa=0.95$, and $\eps=10^{-7}$.

\begin{table}[t]
\begin{minipage}[t]{.48\textwidth}
\begin{algorithm2e}[H]
\small
\caption{Idealized Muon}\label{alg:ideal}
Initialize parameter $W_0$ and buffer $M_{-1}=0$\;
\For{$t=0,1,\ldots$}{
    \tcp{Compute the gradient}
    $G_t\gets\nabla\cL(W_t)$\;
    \tcp{Update the momentum buffer}
    $M_t\gets\beta M_{t-1}+(1-\beta)G_t$\;
    \tcp{Compute a compact SVD}
    $U_t\Sigma_tV_t^\top\gets\operatorname{SVD}(M_t)$\;
    \tcp{Set positive singular values to one}
    $D_t\gets U_tV_t^\top$ (i.e., $D_t\gets\polar(M_t)$)\;
    \tcp{Take an update step}
    $W_{t+1}\gets W_t-\eta_tD_t$\;
}
\end{algorithm2e}
\end{minipage}\hfill
\begin{minipage}[t]{.49\textwidth}
\begin{algorithm2e}[H]
\small
\caption{Muon with Newton–Schulz}\label{alg:practical}
Initialize parameter $W_0$ and buffer $M_{-1}=0$\;
Choose NS polynomial $p$, iteration count $K$\; Choose coefficients $0\leq\beta<1$ and $0\leq\kappa\leq1$\;
\For{$t=0,1,\ldots$}{
    \tcp{Compute the gradient}
    $G_t\gets\nabla\cL(W_t)$\;
    \tcp{Update the momentum buffer}
    $M_t\gets\beta M_{t-1}+(1-\beta)G_t$\;
    $N_t\gets \kappa M_t + (1-\kappa)G_t$\;
    \tcp{Normalize with stabilization}
    $\widehat N_t\gets N_t/(\norm{N_t}_{\F}+\eps)$\;
    \tcp{Apply $K$ iterations and update}
    $W_{t+1}\gets W_t-\eta_t\mathcal P_K(\widehat N_t)$\;
}
\end{algorithm2e}
\end{minipage}
\caption{Idealized and practical Muon for matrix parameters. The idealized update uses an SVD polar factor. The practical update normalizes Nesterov-style momentum and applies $K$ Newton--Schulz iterations. 
% The reference implementation uses the quintic in~\cref{eq:practical-polynomial} with $K=5$ \citep{jordan2024muon}.
}\label{tab:algorithms}
\end{table}
Regarding the polynomial part, \citet{jordan2024muon}
uses the quintic Newton--Schulz polynomial
\begin{equation}\label{eq:practical-polynomial}
p_{\mathrm{quin}}(s)
=3.4445s-4.7750s^3+2.0315s^5,
\end{equation}
with a fixed number of iterations.
We also consider the  cubic Newton--Schulz polynomial
\begin{equation}\label{equ:cubic}
p_{\mathrm{cub}}(s)=\frac32s-\frac12s^3.
\end{equation}
Unlike the quintic polynomial, the  cubic satisfies
$p_{\mathrm{cub}}^{\circ K}(s)\to1$ as $K\to\infty$
for every $s\in(0,1]$.
Consequently, its matrix iteration converges to the exact
polar factor, recovering the orthogonalization used in
idealized Muon (\cref{alg:ideal}). More generally, our analysis allows any
odd polynomial $p$ satisfying the following assumption.
\begin{assumption}[Positive polynomial map]
\label{ass:poly}
The polynomial $p$ is odd, and for the chosen finite $K$,
$(p^{\circ K})'(0)>0$ and
$p^{\circ K}(s)>0$ for all $0<s\leq1$.
\end{assumption}
Both the cubic and quintic satisfy \cref{ass:poly} for every finite $K$. Unless stated otherwise, we consider a fixed finite number $K$ of Newton–Schulz iterations.

\section{Frobenius Max-Margin Bias of Practical Muon}\label{sec:main}

In this section, we first study the implicit bias of practical Muon with a fixed positive stabilizer and a fixed finite number of Newton–Schulz iterations.
Under suitable stepsize conditions, we show that its iterates converge in direction to the Frobenius max-margin classifier, matching the implicit bias of gradient descent.
We establish this result for the momentum-free algorithm under logistic and exponential losses, and then extend the logistic-loss result to EMA and Nesterov-style momentum.

\subsection{Frobenius Max-Margin of practical Muon}\label{sec:frobenius-momentum}

In this subsection, we first consider the momentum-free variant of \cref{alg:practical}, obtained by setting $\beta=0$.
The following theorem establishes Frobenius maximum-margin selection under explicit stepsize conditions, with proof deferred to \cref{app:frobenius}.

\begin{theorem}[Frobenius max-margin bias]
\label{thm:frobenius}
Suppose \cref{ass:separable} holds.
Fix a polynomial $p$ and a finite iteration count $K$
satisfying \cref{ass:poly}, and a stabilizer $\eps>0$.

Define $u_K=\sup_{0<s\leq1}\frac{p^{\circ K}(s)}{s}$ and $R=\max_{i\in[n]}\norm{A_i}_{\F}$.
Let $\gamma>0$ be as in \cref{ass:separable}.
Starting from an arbitrary initialization $W_0\in\R^{d_1\times d_2}$, run \cref{alg:practical} with $\beta=0$ and a constant stepsize $\eta>0$ satisfying the corresponding condition: (i) logistic loss with $\eta\leq4\eps/(R^2u_K)$;
(ii) exponential loss with $\eta R^2u_K e^{\eta Ru_K}\leq\gamma$.
Then:
\begin{equation}
\label{eq:mainbias}
    \lim_{t\to\infty}\cL(W_t)=0,
    \qquad
    \lim_{t\to\infty}\frac{W_t}{\norm{W_t}_{\F}}
    =
    \frac{\widehat W_{\F}}{\norm{\widehat W_{\F}}_{\F}}.
\end{equation}
\end{theorem}
The theorem shows that the loss converges to zero
and the iterates converge in direction to the
Frobenius max-margin separator.
This result can also be extended to EMA and Nesterov-style momentum with logistic loss, and we provide the result below with proof deferred to  \cref{app:general-momentum}.

\begin{corollary}[Frobenius max-margin bias with momentum]
\label{cor:momentum-general}
Suppose \cref{ass:separable} holds.
Fix an odd polynomial $p$ and a finite iteration count $K$ satisfying \cref{ass:poly}, a stabilizer $\eps>0$, and coefficients $0\leq\beta<1$ and $0\leq\kappa\leq1$.
Let $R$ be as in \cref{thm:frobenius}.
Under logistic loss, run \cref{alg:practical} from an arbitrary initialization $W_0\in\R^{d_1\times d_2}$ with a constant stepsize $\eta>0$ satisfying
% \begin{equation}
% \label{eq:momentum-step}
%     \min_{0\leq s\leq1}\frac{p^{\circ K}(s)}s
%     \geq
%     \frac{(R+\eps)\eta R^2u_K^2}{4\eps^2}
%     \left(1+\frac{2\kappa\beta}{1-\beta}\right).
% \end{equation}
\begin{equation}
\label{eq:momentum-step}
\inf_{0<s\leq1}\frac{p^{\circ K}(s)}{s}
\geq
\frac{(R+\eps)\eta R^2u_K^2}{4\eps^2}
\left(1+\frac{2\kappa\beta}{1-\beta}\right).
\end{equation}
Then the loss converges to zero, and the iterates converge in direction to the Frobenius max-margin separator, i.e., \cref{eq:mainbias} holds.
\end{corollary}
\begin{table}[!t]
\centering\small
\setlength{\tabcolsep}{4pt}
\renewcommand{\arraystretch}{1.10}
\caption{
Comparison of implicit-bias behaviors.
Frobenius checkmarks indicate max-margin direction convergence.
Spectral checkmarks indicate spectral max-margin convergence, while
Other indicates counterexamples where the direction
differs from both or the relevant directional limit
fails to exist.
}
\label{tab:mechanism}
\begin{tabular}{@{}llccc@{}}
\toprule
Setting & Polynomial & Frobenius & Spectral & Other\\
\midrule
\multicolumn{5}{@{}l}{\textbf{\textit{Previous results}}}\\
\midrule
Gradient descent
& -- & $\greencheck$ &&\\
Idealized Muon with exact-polar
& -- && $\greencheck$ &\\
\midrule
\multicolumn{5}{@{}l}{\textbf{\textit{Our results on practical Muon}}}\\
\midrule
\multicolumn{5}{@{}l}{\textit{Frobenius Max-Margin Bias}}\\
Fixed $K\geq1$, $\eps>0$ (\cref{thm:frobenius,cor:momentum-general})
& General(\cref{ass:poly}) & $\greencheck$ &&\\
\midrule
\multicolumn{5}{@{}l}{\textit{Disentangling Implementation Choices}}\\
Fixed $K\geq1$, $\eps=0$ (\cref{thm:limits}(i))
& Cubic &&& $\greencheck$\\
& Quintic &&& $\greencheck$\\
$K\to\infty$, $\eps=0$ (\cref{thm:limits}(ii))
& Cubic && $\greencheck$ &\\
& Quintic &&& $\greencheck$\\
$K\to\infty$, fixed $\eps>0$ (\cref{thm:limits}(iii))
& Cubic && $\greencheck$ &\\
\midrule
\multicolumn{5}{@{}l}{\textit{Spectral Max-Margin Recovery}}\\
Growing $K_t$, fixed $\eps>0$ (\cref{thm:recovery})
& Cubic && $\greencheck$ &\\
Growing $K_t$, varying $\eps_t$ (\cref{thm:quintic-recovery})
& Quintic && $\greencheck$ &\\
\bottomrule
\end{tabular}
\end{table}
\subsection{Proof Outline of Theorem~\ref{thm:frobenius}}
We prove \cref{thm:frobenius} by comparing the Muon iterates with gradient descent.
The following lemma shows that the finite Newton–Schulz update with positive stabilization becomes approximately linear as its input approaches zero.
\begin{lemma}[Local linearization of the stabilized update]
\label{lem:taylor}
For fixed finite $K$ and $\eps>0$, let $a_K=(p^{\circ K})'(0)$.
As $\norm{G}_{\F}\to0$ with $G\in\R^{d_1\times d_2}$, we have
\begin{equation}
\label{eq:linearization}
    \mathcal P_K\!\left(\frac{G}{\norm G_{\F}+\eps}\right)
    =
    \frac{a_K}{\eps}G
    +O(\norm G_{\F}^2).
\end{equation}
The implicit constant depends only on $p$, $K$, and $\eps$,
and the remainder bound holds uniformly over bounded sets
of inputs.
\end{lemma}

To control how the approximation errors accumulate, we use the following stability lemma.

\begin{lemma}[Trajectory stability under summable perturbations]
\label{lem:tracking}
Let $f$ be differentiable and fix $\alpha>0$.
Consider the iterations
$W_{t+1}=W_t-\alpha\nabla f(W_t)+r_t$
and
$V_{t+1}=V_t-\alpha\nabla f(V_t)$
for $t\geq T$, with $V_T=W_T$.
Assume that both trajectories remain in a set on which the map
$X\mapsto X-\alpha\nabla f(X)$ is nonexpansive, and that
$\sum_{t=T}^{\infty}\norm{r_t}_{\F}<\infty$.
Then
\begin{equation}
\label{eq:bounded-tracking}
    \sup_{t\geq T}\norm{W_t-V_t}_{\F}
    \leq
    \sum_{t=T}^{\infty}\norm{r_t}_{\F}
    <\infty.
\end{equation}
If $V_t=\widehat W\log t+O(\log\log t)$ for some nonzero $\widehat W$, then the same expansion holds for $W_t$.
In particular, both normalized trajectories converge to
$\widehat W/\norm{\widehat W}_{\F}$.
\end{lemma}

The proofs of \cref{lem:taylor,lem:tracking} are given in \cref{app:taylor,app:tracking-proof}, respectively.

\paragraph{Proof sketch of \cref{thm:frobenius}.}
Bounds on the polynomial update and the stated stepsize conditions yield loss descent and $\sum_t\norm{G_t}_{\F}^2<\infty$.
Thus $G_t\to0$, and separability gives $\cL(W_t)\to0$.
By \cref{lem:taylor}, the update becomes
$W_{t+1}=W_t-\alpha G_t+r_t$,
where $\alpha=\eta a_K/\eps$ and
$\norm{r_t}_{\F}=O(\norm{G_t}_{\F}^2)$.
Consequently, $\sum_t\norm{r_t}_{\F}<\infty$.

% First, the sector estimates for the polynomial update give loss descent and $\sum_t\norm{G_t}_{\F}^2<\infty$. Separability then implies $\lim_{t\to\infty}\cL(W_t)=0$.

% Second, \cref{lem:taylor} writes the Muon trajectory as $W_{t+1}=W_t-\alpha G_t+r_t$, where $\alpha=\eta a_K/\eps$ and $\sum_t\norm{r_t}_{\F}<\infty$. Thus finite iterations and positive stabilization produce a summable perturbation of gradient descent.
We compare this iteration with gradient descent using stepsize $\alpha$, initialized at $V_T=W_T$.
For logistic loss, the global Hessian bound $R^2/4$ and the stepsize condition make the gradient-descent map globally nonexpansive, allowing $T=0$.
For exponential loss, $\norm{\nabla^2\cL(W)}_{\op}\leq R^2\cL(W)$, so we choose $T$ with sufficiently small $\cL(W_T)$.
The established Muon descent and a small-loss descent estimate for gradient descent keep both trajectories in $\{\cL\leq\cL(W_T)\}$, where the gradient-descent map is nonexpansive.

By \cref{lem:tracking}, $\sup_{t\geq T}\norm{W_t-V_t}_{\F}<\infty$.
The classical gradient-descent result of \citet[Theorem~3]{soudry2018implicit} gives
$V_t=\widehat W_{\F}\log t+O(\log\log t)$.
For exponential loss, we apply this result to a globally smooth extension of the scalar loss that leaves the reference trajectory unchanged.
Bounded tracking transfers the same expansion to $W_t$, yielding \cref{eq:mainbias}. \Cref{app:logistic,app:exponential} give the loss-specific details.

\section{Implicit Bias Regimes and Spectral Recovery}\label{sec:spectral}
In the previous section, we established Frobenius max-margin bias for Muon with fixed finite Newton--Schulz iterations and positive stabilization. However, the individual roles of these two implementation choices remain unclear. In this section, we disentangle their effects, examine the role of polynomial choice, and investigate how spectral max-margin behavior can be recovered.

\subsection{Disentangling Finite Orthogonalization and Stabilization}
\label{sec:limits}

For fixed finite $K$ and $\eps>0$, \cref{thm:frobenius}
establishes Frobenius max-margin selection on general
separable data. We now distinguish the roles of iteration
count, stabilization, and polynomial choice.
\Cref{tab:mechanism} summarizes the resulting guarantees
and counterexamples. The following theorem characterizes these regimes,
with proofs in \cref{app:limits}.

\begin{theorem}
\label{thm:limits}
Under \cref{ass:separable}, consider
\cref{alg:practical} with $\beta=0$
and the empirical loss in
\cref{equ:empiricalloss} with either
logistic or exponential loss.
The following statements hold.

\begin{enumerate}[label=(\roman*),leftmargin=1.4em,
itemsep=3pt,topsep=4pt]

\item Fixed $K\geq1$, $\eps=0$, for any nonzero signed feature matrix $A=yX$, any positive stepsize sequence $(\eta_t)_{t\geq0}$, initial weight $W_0=0$, and a polynomial satisfying \cref{ass:poly}, define $D_K=\mathcal P_K(A/\norm{A}_{\F})$. Then, we have: \begin{equation}\label{eq:finite-ns-direction}\lim_{t\to\infty}\frac{W_t}{\norm{W_t}_{\F}}=\frac{D_K}{\norm{D_K}_{\F}}.\end{equation}For either the cubic or the quintic and every finite $K\geq1$, there exists an $A$ such that the selected direction \textbf{differs from both Frobenius and spectral max-margin directions}.
\item $K\to\infty$, $\eps=0$. For the cubic
with any finite $W_0$ and
$\eta_t=\eta_0/(t+1)$ for $\eta_0>0$,
the spectral max-margin is achieved:
\begin{equation}\label{eq:iteration-general}
\lim_{T\to\infty}\limsup_{K\to\infty}
\left(
\gamma_{\op}
-
\frac{\min_i\ip{A_i}{W_T^{(K)}}}
{\norm{W_T^{(K)}}_{\op}}
\right)=0,
\end{equation}
where $W_T^{(K)}$ denotes the iterate obtained
using the same finite $K$ at every update.

In contrast, for the quintic, there exists a counterexample such that, for any positive stepsize sequence, the normalized iterates \textbf{fail to converge} as $K\to\infty$ at every fixed $T\geq1$.

\item $K\to\infty$, fixed $\eps>0$.
For the cubic, under \cref{ass:separable},
with any finite $W_0$ and the harmonic
stepsize as in part (ii), the spectral-margin
guarantee \cref{eq:iteration-general}
continues to hold for every fixed $\eps>0$.

\end{enumerate}
\end{theorem}

The results in \cref{thm:limits} disentangle the roles of stabilization, iteration count, and polynomial choice in determining implicit bias. Part (i) shows that removing stabilization ($\eps=0$) can eliminate the Frobenius max-margin bias established in \cref{thm:frobenius}, without necessarily recovering the spectral max-margin bias of idealized Muon. With finite Newton--Schulz iterations, the selected direction can differ from both max-margin directions. Part (ii) shows that as $K\to\infty$ with $\eps=0$, spectral max-margin can be recovered for the cubic polynomial, consistent with $p_{\mathrm{cub}}^{\circ K}(s)\to1$ for every $s\in(0,1]$. In contrast, the quintic does not share this convergence property, and its normalized iterates can fail to converge as $K\to\infty$. Finally, part (iii) shows that for the cubic polynomial, orthogonalization plays a dominant role in determining implicit bias as $K\to\infty$: the spectral max-margin is achieved regardless of the fixed stabilization level $\eps\geq0$.

 To illustrate the key idea behind the counterexamples, consider a single-example problem with $A=\diag(2,1)$, $W_0=0$, and either logistic or exponential loss. By a simple calculation, the Frobenius and spectral hard-margin solutions are $\widehat W_{\F}=A/5$ and $\widehat W_{\op}=\Id_2/3$, respectively, with diagonal-entry ratios $2$ and $1$. When $\eps=0$, the cubic polynomial satisfies $1<(D_K)_{11}/(D_K)_{22}<2$ for every finite $K\geq1$, while the quintic polynomial with $K=5$ satisfies $5/9<(D_5)_{11}/(D_5)_{22}<3/4$. By \cref{eq:finite-ns-direction}, both select directions that achieve neither Frobenius nor spectral max-margin.

\subsection{Recovering Spectral Max-Margin}
\label{sec:recovery}

The previous subsection shows that fixed finite
Newton--Schulz iterations do not necessarily recover
the spectral max-margin direction, whereas
taking $K\to\infty$ can recover the spectral margin
under suitable conditions.
However, infinite Newton--Schulz iterations
are computationally infeasible in practice.
This motivates the question of whether the spectral
max-margin bias of idealized Muon can be
recovered using finite Newton--Schulz
iterations at each training step.
We address this question by allowing the iteration
count $K_t$ to grow with training time. We first establish spectral max-margin recovery
for the cubic polynomial with fixed $\eps>0$.
\begin{theorem}[Spectral recovery for cubic polynomial]
\label{thm:recovery}
Under \cref{ass:separable}, consider
\cref{equ:empiricalloss} with exponential loss.
Run \cref{alg:practical} with the cubic polynomial,
$\beta=0$, any finite $W_0$, and fixed $\eps>0$.
Define $R_*=\max_i\norm{A_i}_*$.
Choose $0<\eta_0\leq1/R_*$, stepsize
$\eta_t=\eta_0/(t+1)$, and $
K_t=\left\lceil
\frac{2\log(t+2)}{\log(3/2)}
\right\rceil.$ Then $\lim_{T\to\infty}\cL(W_T)=0$ and
\begin{equation}\label{eq:recovery-margin}
0\leq
\gamma_{\op}
-\frac{\min_i\ip{A_i}{W_T}}{\norm{W_T}_{\op}}
=O\!\left(\frac{1}{\log(T+1)}\right)
\quad\text{as }T\to\infty.
\end{equation}
\end{theorem}

\paragraph{Proof sketch.}
Let $G_t=\nabla\cL(W_t)$ denote the gradient
at iteration $t$.
Recall from \cref{eq:spectral-oracle} that
the exact polar factor achieves the optimal
gradient alignment
$\norm{G_t}_*=\ip{G_t}{\polar(G_t)}$.
For $G_t\ne0$, the relative gradient-alignment
gap of the finite Newton--Schulz update satisfies
(see \cref{lem:dual-oracle-recovery})
\begin{equation}\label{eq:weighted-defect}
0\leq
\frac{
\norm{G_t}_*-
\ip{G_t}{
\mathcal P_{K_t}\!\left(
\frac{G_t}{\norm{G_t}_{\F}+\eps}
\right)
}
}{
\norm{G_t}_*
}
\leq
r\left(1+\frac{\eps}{\norm{G_t}_{\F}}\right)
(2/3)^{K_t},
\end{equation}
where $r=\min\{d_1,d_2\}$. The bound shows that the approximation error
decreases exponentially with $K$, while the
factor $\eps/\norm G_{\F}$ grows as the
gradient vanishes. To control this factor, the bounded operator
norm of the cubic update gives
$\norm{W_t}_{\op}
\leq\norm{W_0}_{\op}+\sum_{s<t}\eta_s$.
Combining this bound with the positive-margin condition in \cref{ass:separable} yields $
\norm{G_t}_{\F}
\geq
\frac{\gamma e^{-1-R_*\norm{W_0}_{\op}}}{t+1}.$ Substituting this lower bound and the chosen
$K_t$ into \cref{eq:weighted-defect} bounds
the relative oracle error by $C/(t+2)$,
where $C$ depends on the data, $W_0$,
and the fixed stabilization constant.
Consequently, its stepsize-weighted sum
is finite, whereas
$\sum_{t<T}\eta_t$ grows logarithmically.
Applying the inexact spectral-margin bound in
\cref{lem:inexact-spectral-recovery}
then yields \cref{eq:recovery-margin}. The full proof is given in
\cref{app:recovery-proof}.

\vspace{2mm}

We next consider spectral max-margin recovery for the
original quintic polynomial. Unlike the cubic, whose
Newton--Schulz iterates converge to the exact polar
factor as $K\to\infty$, the quintic does not generally
have this convergence property, as discussed in
\cref{sec:limits}. Consequently, increasing the iteration
count alone does not necessarily provide the same recovery mechanism
as in the cubic case. We instead establish spectral
max-margin recovery by allowing both the iteration count
and the stabilization parameter to vary over training.
The following theorem gives a guarantee
under a gradient-dependent stabilization
scheme, with proof deferred to \cref{app:quintic-recovery}.
\begin{theorem}[Spectral recovery for quintic polynomial]\label{thm:quintic-recovery}
Under \cref{ass:separable}, run \cref{alg:practical} with exponential loss, the quintic in~\cref{eq:practical-polynomial}, $\beta=0$, and any finite $W_0$. Choose stepsize $\eta_t=\eta_0/(t+1)$ for any $\eta_0>0$. Set $K_t=\left\lceil\frac{\log(t+2)}{\log p_{\mathrm{quin}}^{\prime}(0)}\right\rceil$ and $\eps_t=\norm{G_t}_{\F}\bigl((t+2)^{U_t}-1\bigr)$, where $U_t\sim\operatorname{Unif}(0,1)$. Then, almost surely, $\lim_{T\to\infty}\cL(W_T)=0$ and
\begin{equation}\label{eq:quintic-random-margin}
0\leq
\gamma_{\op}
-\frac{\min_i\ip{A_i}{W_T}}{\norm{W_T}_{\op}}
=
O\!\left(
\frac{\log\log(T+2)}{\log(T+1)}
\right)
\quad\text{as }T\to\infty.
\end{equation}

\end{theorem}
\begin{remark}
The scheme of $\epsilon_t$ in \cref{thm:quintic-recovery}
depends on the gradient norm and an independent random
variable, unlike the fixed positive $\epsilon$ used
in practical Muon and previous theorems. This limits the theorem's direct
applicability to the standard implementation, although
it establishes that spectral max-margin recovery is
possible for the original quintic polynomial under
a suitable scheme. Whether spectral
max-margin can be recovered for the quintic with
fixed $\eps>0$ remains open.
\end{remark}

\section{Experiments}\label{sec:experiments}
\begin{figure}[t]
\centering
\includegraphics[width=\textwidth]{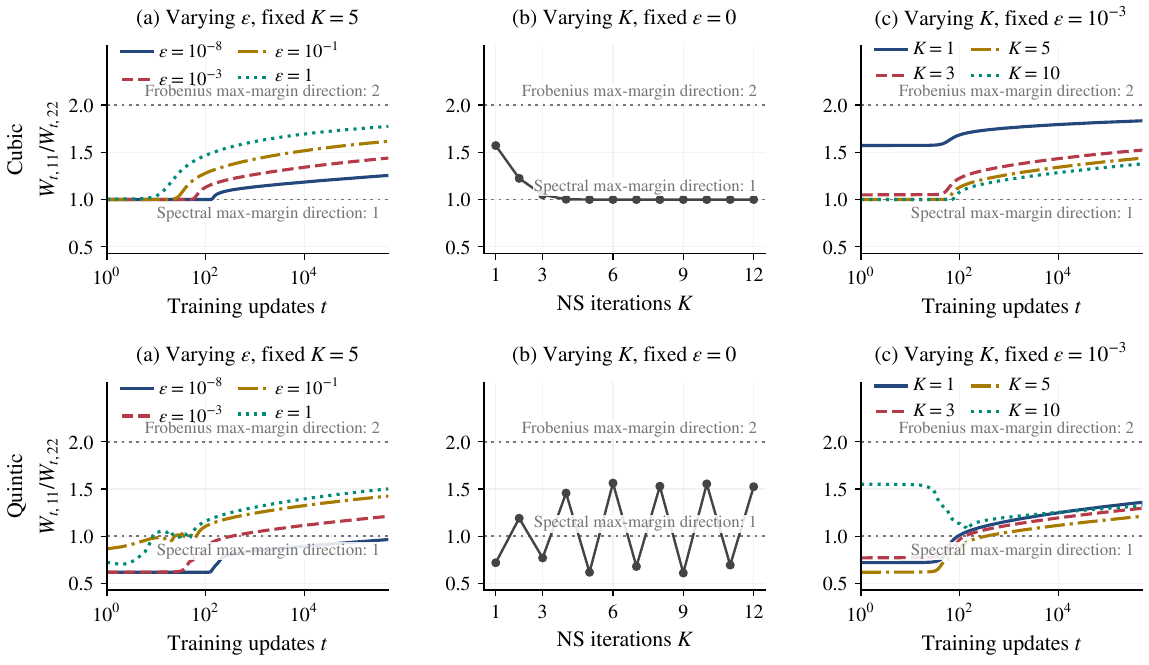}
\caption{
Muon training using cubic (top) and quintic (bottom)
Newton--Schulz iterations following the setting of \cref{sec:single-experiment}. We track the predictor
ratio $W_{t,11}/W_{t,22}$, with reference values
$(\widehat W_{\F})_{11}/(\widehat W_{\F})_{22}=2$
and
$(\widehat W_{\op})_{11}/(\widehat W_{\op})_{22}=1$
for the Frobenius and spectral max-margin directions,
respectively.
}
\label{fig:single}
\end{figure}
We empirically investigate the implicit-bias behaviors
established in previous sections.
We train the classifiers
of \cref{equ:empiricalloss} with exponential loss. The experiments here are momentum-free version of \cref{alg:practical}, with
additional momentum experiments provided in
\cref{app:experiments}.

\subsection{Finite iterations and stabilization}\label{sec:single-experiment}
Following the proof sketch of \cref{thm:limits},
we first use the same single-example dataset
with signed feature $A=\diag(2,1)$.
We track the predictor ratio $W_{t,11}/W_{t,22}$
and compare it with the corresponding ratios
of the Frobenius and spectral max-margin directions,
$(\widehat W_{\F})_{11}/(\widehat W_{\F})_{22}=2$
and
$(\widehat W_{\op})_{11}/(\widehat W_{\op})_{22}=1$,
respectively. We initialize $W_0=0$ and run the cubic and quintic
variants with constant stepsize $\eta=0.05$ for
$5\times10^5$ updates. \Cref{fig:single} compares the cubic and quintic
polynomials in its top and bottom rows, respectively. Specifically:
\begin{itemize}
    \item Column (a) fixes $K=5$ and varies positive $\eps$.
For both polynomials, the predictor ratios approach
the Frobenius max-margin target as training progresses,
consistent with \cref{thm:frobenius}.
Interestingly, larger $\eps$ leads to an earlier
transition toward the Frobenius max-margin direction.
\item Column (b) sets $\eps=0$, where the predictor ratio
remains constant throughout training for each fixed
$K$, as established in \cref{thm:limits}(i).
As $K$ increases, the cubic approaches the spectral
max-margin ratio, consistent with
\cref{thm:limits}(ii), whereas the quintic exhibits
nonmonotonic behavior, consistent with the
result in \cref{thm:limits}(ii).
\item Column (c) fixes $\eps=10^{-3}$ and varies $K$.
For both polynomials, the predictor ratios approach
the Frobenius max-margin target as training progresses,
consistent with \cref{thm:frobenius}.
The iteration count $K$ affects the transient trajectories and the approach to the Frobenius max-margin direction, without changing the asymptotic implicit bias.
\end{itemize}
\subsection{Frobenius selection and spectral recovery}\label{sec:matrix-experiments}\label{sec:recovery-experiment}
We next generate five synthetic datasets using
independent random seeds, each with $n=12$
and $d_1=d_2=4$. The datasets are constructed
to have known Frobenius hard-margin solutions,
as detailed in \cref{app:matrix-data}. We compare three methods: Muon
(\cref{alg:practical}) with cubic Newton--Schulz
iterations with fixed $K=3$, with growing
$K_t=\left\lceil 2\log(t+2)/\log(3/2)\right\rceil$, and idealized Muon
(\cref{alg:ideal}) with exact polar orthogonalization.
All methods use $\beta=0$, start from $W_0=0$,
and run for $3\times10^5$ updates with the same stepsize $\eta_t=0.5/(t+1)$.
Both practical Muon variants use fixed $\eps=0.1$. We evaluate the normalized minimum margins
$\min_i\ip{A_i}{W_t}/\norm{W_t}_{\F}$ and
$\min_i\ip{A_i}{W_t}/\norm{W_t}_{\op}$,
relative to their respective optimal values
$\gamma_{\F}$ and $\gamma_{\op}$ defined in
\cref{eq:spectral-margin}.
A value of one indicates that the corresponding
max-margin is achieved.
We also measure the relative spectral-oracle error
$1-\ip{G_t}{D_t}/\norm{G_t}_*$,
where $G_t=\nabla\cL(W_t)$ and
$D_t=(W_t-W_{t+1})/\eta_t$.
This quantity measures the fraction of optimal
gradient alignment lost by the computed update,
as bounded in \cref{eq:weighted-defect}. All metrics are averaged across the five datasets.

\begin{figure}[t]
\centering
\includegraphics[width=0.886484\textwidth]{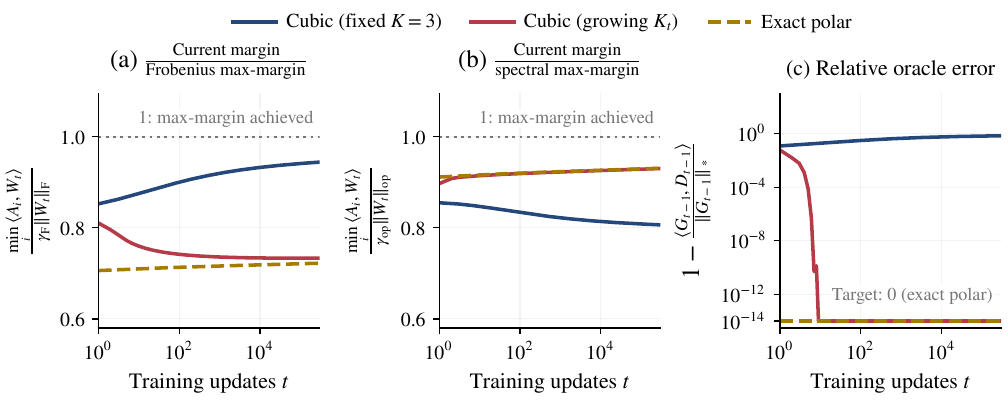}
\vspace{-3mm}
\caption{Frobenius max-margin selection and spectral recovery
under the experimental settings described in
\cref{sec:matrix-experiments}.
Results are averaged over five random seeds.}\label{fig:matrix}
\end{figure}

\Cref{fig:matrix}(a) shows that the fixed-$K$
cubic variant favors the Frobenius max-margin,
consistent with \cref{thm:frobenius}.
In contrast, panels (b,c) show that the growing-$K_t$
cubic variant closely tracks idealized Muon in
spectral margin and oracle accuracy, consistent
with \cref{thm:recovery}.
The fixed-$K$ variant exhibits a lower spectral
margin and a larger oracle error.

\section{Related Work}

\subsection{Implicit bias of optimization}
Gradient descent and stochastic gradient descent select the Euclidean max-margin direction on separable linear problems under suitable loss and stepsize conditions \citep{soudry2018implicit,nacson2019stochastic}. Subsequent work connects this behavior to loss tails, regularization paths, and primal-dual geometry \citep{nacson2019convergence,ji2020regularization,ji2021primaldual}. Beyond Euclidean geometry, steepest descent, mirror descent, and mirror flow exhibit max-margin bias governed by the optimization geometry \citep{gunasekar2018geometry,sun2023unified,wang2023faster,pesme2024mirror}. For homogeneous neural networks, analyses of gradient flow and steepest descent establish directional convergence and margin-stationarity results \citep{lyu2020margin,ji2020directional,tsilivis2025flavors}.

For matrix-valued parameters, \citet{fan2025spectral} establish spectral max-margin behavior for spectral descent and Muon with exact orthogonalization on multiclass separable data. Related work examines the effects of momentum in homogeneous networks and separable classification, as well as sampling and variance reduction in stochastic steepest descent \citep{gronich2026homogeneous,wang2022momentum,li2026minibatch}. Studies of AdaGrad and Adam further examine the implicit bias of adaptive updates, including the role of stabilization \citep{qian2019adagrad,wang2021adaptive,zhang2024adam}. More recently, \citet{wang2026annealing} study how annealed stabilization affects the implicit bias of smoothed-sign descent.
\subsection{Steepest descent}
\citet{kelner2013flow} develop a non-Euclidean generalization of gradient descent under arbitrary norms and apply it to approximate maximum-flow problems. In neural-network optimization, \citet{carlson2015stochastic} introduce stochastic spectral descent for restricted Boltzmann machines, using the spectral norm to exploit the matrix structure of the objective. \citet{carlson2015preconditioned} combine non-Euclidean descent with preconditioning and study restricted Boltzmann machines, feedforward networks, and convolutional networks. \citet{bernstein2024norm} interpret several modern optimizers through steepest descent under parameter-dependent norms, including operator norms for matrix weights. Muon uses a small number of Newton--Schulz iterations to approximate orthogonalized momentum updates and has been applied to language-model and image-classification training \citep{jordan2024muon}. \citet{liu2025scalable} scale Muon to large language models through weight decay, update scaling, and distributed computation. Polynomial iterations have also been optimized for fast polar approximation in Muon \citep{amsel2025polar}.

\citet{kim2026} establish convergence guarantees for Muon with Newton--Schulz iterations, whereas \citet{oliveira2026convergence} analyze a soft-sign surrogate. \citet{li2026smoothing} investigate the smoothing effects of finite Newton--Schulz iterations in nonsmooth nonconvex optimization, while \citet{peng2026practical} analyze convergence of practical Muon with finite Newton--Schulz iterations and Nesterov-style momentum. Unlike these convergence analyses, our work focuses on how the Newton--Schulz iteration count, stabilization constant, and polynomial choice determine the asymptotic max-margin behavior.

\section{Conclusion}
This paper establishes that Muon with finite Newton--Schulz iterations and positive stabilization selects the Frobenius max-margin direction in separable linear classification. We also characterize the effects of stabilization, iteration count, and polynomial choice, and provide conditions for recovering spectral max-margin behavior. Our results focus on full-batch linear classification, leaving stochastic optimization and nonlinear neural networks beyond the present analysis. Motivated by the asymptotic connection to gradient descent in linear classification, an interesting direction for future work is to investigate whether similar behavior emerges in later stages of large-scale neural network training.

% Authors are strongly encouraged to include an anonymous AI Disclosure section
% if generative AI was used substantively. Uncomment and replace the example:
% \section*{AI Disclosure}
% We used [tool] to [briefly explain how it was used].

% Acknowledgments---Will not appear in anonymized version
% \acks{We thank our colleagues and funding agencies;
% automatically hides this text.}
\bibliography{references}
\newpage
\appendix
\section*{Contents of the Appendix}
The Appendix is organized as follows:
\begin{itemize}[nosep]
    \item \cref{app:notation} summarizes the main notation used in this work.
    \item \Cref{app:frobenius} contains the proofs for \cref{sec:main}.
    \item \Cref{app:limits} gives auxiliary lemmas and separate proofs of \cref{thm:limits}(i)--(iii).
    \item \Cref{app:recovery} gives auxiliary lemmas and proofs of \cref{thm:recovery,thm:quintic-recovery}.
    \item \Cref{app:experiments} provides additional experimental details and supplementary results.
\end{itemize}

\section{Notation}
\label{app:notation}
Table~\ref{tab:notation} summarizes the notation used in the main text and proofs.

\begin{table}[h]
\centering
\small
\setlength{\tabcolsep}{5pt}
\renewcommand{\arraystretch}{1.12}
\begin{threeparttable}
\caption{Core symbols and constants used in this paper.}
\label{tab:notation}
\begin{tabular}{@{}p{3.1cm}p{2.0cm}p{8.0cm}@{}}
\toprule
Symbol & Type & Description \\
\midrule
$A_i$, $W_t$ & Matrices & Signed matrix feature and parameter at training iteration $t$, both in $\R^{d_1\times d_2}$. \\[2pt]
$\cL$, $G_t$ & Loss, gradient & Empirical loss and $G_t=\nabla\cL(W_t)$. \\[2pt]
$\ip A B$ & Inner product & Frobenius inner product $\operatorname{tr}(A^\top B)$. \\[2pt]
$\norm{\cdot}_{\F}$, $\norm{\cdot}_{\op}$, $\norm{\cdot}_*$ & Matrix norms & Frobenius, spectral, and nuclear norms. \\[2pt]
$R$, $R_*$, $\gamma$ & $>0$ & Feature bounds $R=\max_i\norm{A_i}_{\F}$ and $R_*=\max_i\norm{A_i}_*$, and a separating margin $0<\gamma\leq\gamma_{\F}$ from \cref{ass:separable}. \\[2pt]
$H$ & $>0$ & Local Hessian bound in the auxiliary lemmas and proofs, specified where it is used. \\[2pt]
$p$, $p^{\circ K}$ & Scalar maps & Scalar polynomial iteration and its $K$-fold composition. \\[2pt]
$\mathcal P_K$ & Matrix map & Spectral polynomial map $\mathcal P_K(G)=Up^{\circ K}(\Sigma)V^\top$ for a compact SVD $G=U\Sigma V^\top$. \\[2pt]
$K$, $\eps$ & Parameters & Newton--Schulz iteration count and nonnegative stabilizer, with $K_t$ and $\eps_t$ denoting time-dependent choices. \\[2pt]
$a_K$, $l_K$, $u_K$ & $>0$ & Slope $(p^{\circ K})'(0)$ and lower and upper sector bounds for $p^{\circ K}(s)/s$ on $[0,1]$. \\[2pt]
$\eta$, $\eta_t$, $\alpha$ & $>0$ & Constant or time-dependent outer stepsize, and effective gradient-descent step $\alpha=\eta a_K/\eps$ for fixed $K$ and $\eps>0$. \\[2pt]
$M_t$, $N_t$ & Matrices & EMA buffer and input to the normalized polynomial map. \\[2pt]
$\beta$, $\kappa$ & Scalars & Momentum coefficient $\beta\in[0,1)$ and input weight $\kappa\in[0,1]$. \\[2pt]
$\widehat W_{\F}$, $\widehat W_{\op}$ & Matrices & Frobenius and spectral hard-margin solutions in \cref{eq:hardmargin,eq:spectral-hardmargin}, with $\widehat W_{\F}$ unique. \\[2pt]
$\gamma_{\F}$, $\gamma_{\op}$ & $>0$ & Frobenius and spectral max-margins, equal to $1/\norm{\widehat W_{\F}}_{\F}$ and $1/\norm{\widehat W_{\op}}_{\op}$, respectively, in \cref{eq:spectral-margin}. \\[2pt]
$r$ & $\mathbb N_{>0}$ & Matrix dimension bound $r=\min\{d_1,d_2\}$. \\
\bottomrule
\end{tabular}
\end{threeparttable}
\end{table}
\section{Proof of Section~\ref{sec:main}}\label{app:frobenius}

We first provide auxiliary lemmas and prove \cref{lem:taylor,lem:tracking}.
We then prove \cref{thm:frobenius} for logistic and exponential losses and establish the momentum extension in \cref{cor:momentum-general}.

\subsection{Auxiliary lemmas}\label{app:spectral}\label{app:core}

Fix a finite iteration count $K$ and a stabilizer $\eps>0$, and define
\begin{equation}\label{eq:sector-constants}
l_K=\min_{0\leq s\leq1}\frac{p^{\circ K}(s)}s,
\end{equation}
with the ratio at zero defined by continuity as $a_K$. Continuity and \cref{ass:poly} give $0<l_K\leq a_K\leq u_K<\infty$. The following lemma bounds the alignment and norm of the polynomial update.

\begin{lemma}[Alignment and norm bounds]
\label{lem:sector}
Suppose \cref{ass:poly} holds.
For every matrix $G$, set $g=\norm{G}_{\F}, P=\mathcal P_K(G/(g+\eps)).$
Then
\begin{equation}
\label{eq:sector-bounds}
    \frac{l_Kg^2}{g+\eps}
    \leq \langle G,P\rangle
    \leq \frac{u_Kg^2}{g+\eps},
    \qquad
    \norm{P}_{\F}^2
    \leq \frac{u_K}{g+\eps}\langle G,P\rangle.
\end{equation}
In particular,
$\norm{P}_{\F}\leq u_Kg/(g+\eps)\leq u_K$.
\end{lemma}

\begin{proof}
The case $G=0$ is immediate.
Otherwise, write $G=U\operatorname{diag}(\sigma_j)V^\top$
as a compact SVD and set $s_j=\sigma_j/(g+\eps)\in(0,1)$.
The polynomial map gives
\[
    P=U\operatorname{diag}\!\bigl(p^{\circ K}(s_j)\bigr)V^\top.
\]
Using $l_Ks_j\leq p^{\circ K}(s_j)\leq u_Ks_j$
and $\sum_j\sigma_j^2=g^2$, we obtain
\[
    \frac{l_Kg^2}{g+\eps}
    \leq \langle G,P\rangle
    =\sum_j\sigma_jp^{\circ K}(s_j)
    \leq \frac{u_Kg^2}{g+\eps}.
\]
Moreover, positivity and the same upper bound imply
\[
    \norm{P}_{\F}^2
    =\sum_j\bigl(p^{\circ K}(s_j)\bigr)^2
    \leq
    \frac{u_K}{g+\eps}
    \sum_j\sigma_jp^{\circ K}(s_j)
    =
    \frac{u_K}{g+\eps}\langle G,P\rangle.
\]
Combining the upper bounds gives
$\norm{P}_{\F}\leq u_Kg/(g+\eps)\leq u_K$.
\end{proof}

Then we verify the nonexpansiveness condition in \cref{lem:tracking}.

\begin{lemma}[Nonexpansive gradient map]
\label{lem:nonexpansive}
Let $\mathcal C$ be a convex set, and let $f$ be twice continuously
differentiable on a neighborhood of $\mathcal C$.
Suppose that, for some $H>0$,
\[
    0\preceq\nabla^2f(Z)\preceq HI
    \qquad\text{for every }Z\in\mathcal C.
\]
Then, for $0<\alpha\leq2/H$ and $X,Y\in\mathcal C$,
\[
    \norm{X-\alpha\nabla f(X)-Y+\alpha\nabla f(Y)}_{\F}
    \leq \norm{X-Y}_{\F}.
\]
\end{lemma}

\begin{proof}
We identify matrices with their vectorizations, under which the
Frobenius norm equals the Euclidean norm.
Define
\[
    B=\int_0^1\nabla^2f\bigl(Y+s(X-Y)\bigr)\,ds.
\]
Convexity of $\mathcal C$ and the Hessian bounds imply that
$B$ is symmetric and $0\preceq B\preceq HI$.
The fundamental theorem of calculus gives
\[
    \nabla f(X)-\nabla f(Y)=B(X-Y).
\]
Since $0<\alpha\leq2/H$, the eigenvalues of $I-\alpha B$
lie in $[1-\alpha H,1]\subseteq[-1,1]$, so
\[
    \norm{I-\alpha B}_{\mathrm{op}}\leq1.
\]
Consequently,
\[
\begin{aligned}
    \norm{X-\alpha\nabla f(X)-Y+\alpha\nabla f(Y)}_{\F}
    &=\norm{(I-\alpha B)(X-Y)}_{\F}
    \\
    &\leq
    \norm{I-\alpha B}_{\mathrm{op}}\norm{X-Y}_{\F}
    \\
    &\leq \norm{X-Y}_{\F}.
\end{aligned}
\]
\end{proof}

Let $R$ be as in \cref{thm:frobenius}, and let $U$ and $\gamma$
be as in \cref{ass:separable}.
Write $m_i(W)=\langle A_i,W\rangle$.
We first consider logistic loss.

\begin{lemma}[Logistic loss and gradient]
\label{lem:logistic-gradient}
For logistic loss, the gradient and Hessian satisfy
\[
    \norm{\nabla\cL(W)}_{\F}\leq R,
    \qquad
    0\preceq\nabla^2\cL(W)\preceq\frac{R^2}{4}I
    \qquad\text{for every }W.
\]
If $\norm{\nabla\cL(W_t)}_{\F}\to0$, then
$m_i(W_t)\to\infty$ for every $i$, and $\cL(W_t)\to0$.
\end{lemma}

\begin{proof}
Set $r_i(W)=(1+e^{m_i(W)})^{-1}$.
Differentiation gives
\[
\begin{aligned}
    \nabla\cL(W)
    &=-\frac1n\sum_i r_i(W)A_i,
    \\
    \nabla^2\cL(W)
    &=\frac1n\sum_i r_i(W)(1-r_i(W))
    \operatorname{vec}(A_i)\operatorname{vec}(A_i)^\top.
\end{aligned}
\]
Since $0<r_i(W)<1$ and $r_i(W)(1-r_i(W))\leq1/4$,
\[
\begin{aligned}
    \norm{\nabla\cL(W)}_{\F}
    &\leq\frac1n\sum_i r_i(W)\norm{A_i}_{\F}
    \leq R,
    \\
    0\preceq\nabla^2\cL(W)
    &\preceq
    \frac1{4n}\sum_i
    \operatorname{vec}(A_i)\operatorname{vec}(A_i)^\top
    \preceq\frac{R^2}{4}I.
\end{aligned}
\]
Separability and $\norm{U}_{\F}=1$ give
\[
    \frac{\gamma}{n}\sum_i r_i(W)
    \leq
    \frac1n\sum_i r_i(W)\langle U,A_i\rangle
    =
    \langle U,-\nabla\cL(W)\rangle
    \leq
    \norm{\nabla\cL(W)}_{\F}.
\]
If $\norm{\nabla\cL(W_t)}_{\F}\to0$, nonnegativity implies
$r_i(W_t)\to0$ for every $i$, hence $m_i(W_t)\to\infty$.
Each term $\log(1+e^{-m_i(W_t)})$ therefore tends to zero,
so $\cL(W_t)\to0$.
\end{proof}

For exponential loss, \cref{ass:separable} and
$\norm{A_i}_{\F}\leq R$ give
\begin{equation}
\label{eq:exp-gradient-bounds}
    \gamma\cL(W)
    \leq\norm{\nabla\cL(W)}_{\F}
    \leq R\cL(W),
    \qquad
    \norm{\nabla^2\cL(W)}_{\mathrm{op}}
    \leq R^2\cL(W).
\end{equation}
Moreover, for any displacement $D$,
\[
    \cL(W+D)\leq e^{R\norm{D}_{\F}}\cL(W).
\]

\begin{lemma}[Exponential gradient descent from small loss]
\label{lem:exp-tail-gd}
Suppose \cref{ass:separable} holds, and consider exponential loss.
Fix $\alpha>0$ and an initial point $V_T$, and set
$\cL_*=\cL(V_T)$.
If $\alpha nR^2\cL_*\leq1/2$, then gradient descent
\[
    V_{t+1}=V_t-\alpha\nabla\cL(V_t),
    \qquad t\geq T,
\]
satisfies $\cL(V_t)\leq\cL_*$ for every $t\geq T$ and
\[
    V_t=\widehat W_F\log t+O(\log\log t)
    \qquad\text{as }t\to\infty.
\]
\end{lemma}

\begin{proof}
Set $m_*=-\log(n\cL_*)$ and define
\begin{equation}
\label{eq:extended-exp-loss}
    \widetilde\ell(u)=
    \begin{cases}
        e^{-u}, & u\geq m_*,\\
        e^{-m_*}\bigl(1-(u-m_*)+(u-m_*)^2/2\bigr),
        & u<m_*.
    \end{cases}
\end{equation}
This function is positive, strictly decreasing, and $C^2$,
with $0<\widetilde\ell''(u)\leq n\cL_*$.
The extended objective
$\widetilde\cL(V)=n^{-1}\sum_i\widetilde\ell(m_i(V))$
therefore satisfies
\[
    0\preceq\nabla^2\widetilde\cL(V)
    \preceq nR^2\cL_* I.
\]

If either objective is at most $\cL_*$, each scalar loss term
is at most $n\cL_*=e^{-m_*}=\widetilde\ell(m_*)$,
so strict monotonicity gives $m_i(V)\geq m_*$ for every $i$.
Consequently,
\[
    \{V:\widetilde\cL(V)\leq\cL_*\}
    =
    \{V:\cL(V)\leq\cL_*\},
\]
and both objectives and their gradients agree on this set.

Whenever $\cL(V_t)\leq\cL_*$, smoothness gives
\begin{equation}
\label{eq:reference-exp-descent}
\begin{aligned}
    \widetilde\cL(V_{t+1})
    &\leq \cL(V_t)
    -\alpha\left(1-\frac{\alpha nR^2\cL_*}{2}\right)
    \norm{\nabla\cL(V_t)}_{\F}^2\\
    &\leq \cL(V_t)\leq\cL_*.
\end{aligned}
\end{equation}
The equality of the sublevel sets implies
$\cL(V_{t+1})=\widetilde\cL(V_{t+1})\leq\cL(V_t)$.
Induction proves the loss bound and shows that the sequence
is also gradient descent on $\widetilde\cL$.

The extension has an exact exponential derivative tail.
The vectorized data matrix has squared spectral norm at most
$nR^2$, and $\alpha nR^2\cL_*\leq1/2<2$.
Thus \citet[Theorem~3]{soudry2018implicit} applies,
with stepsize $\alpha/n$ for the summed loss, yielding
\[
\begin{aligned}
    V_t
    &=\widehat W_F\log(t-T)+O(\log\log(t-T))\\
    &=\widehat W_F\log t+O(\log\log t),
\end{aligned}
\]
where the last equality uses that $T$ is fixed.
\end{proof}

\subsection{Proof of Lemma~\ref{lem:taylor}}
\label{app:taylor}

\begin{proof}
Oddness gives $p^{\circ K}(s)=a_Ks+O(s^3)$ as $s\to0$, so
\[
    b_K
    =\sup_{0<s\leq1}
    \frac{|p^{\circ K}(s)-a_Ks|}{s^3}
    <\infty.
\]
Write $g=\norm{G}_{\F}$ and let $\sigma_j$ be the singular values of $G$.
Applying the scalar remainder bound and using
$\sum_j\sigma_j^6\leq(\sum_j\sigma_j^2)^3=g^6$ yields
\[
    \left\|
        \mathcal P_K\!\left(\frac{G}{g+\eps}\right)
        -\frac{a_KG}{g+\eps}
    \right\|_{\F}
    \leq
    \frac{b_K}{(g+\eps)^3}
    \left(\sum_j\sigma_j^6\right)^{1/2}
    \leq
    \frac{b_Kg^3}{(g+\eps)^3}.
\]
Also,
\[
    \left\|
        \frac{a_KG}{g+\eps}-\frac{a_KG}{\eps}
    \right\|_{\F}
    =
    \frac{a_Kg^2}{\eps(g+\eps)}.
\]
The triangle inequality therefore gives
\begin{equation}
\label{eq:quadratic-remainder}
\begin{aligned}
    \left\|
        \mathcal P_K\!\left(\frac{G}{g+\eps}\right)
        -\frac{a_K}{\eps}G
    \right\|_{\F}
    &\leq
    \frac{b_Kg^3}{(g+\eps)^3}
    +\frac{a_Kg^2}{\eps(g+\eps)}
    \\
    &\leq
    \frac{a_K+b_K}{\eps^2}g^2,
\end{aligned}
\end{equation}
where the last step uses $g/(g+\eps)\leq1$ and $g+\eps\geq\eps$.
\end{proof}

\subsection{Proof of Lemma~\ref{lem:tracking}}
\label{app:tracking-proof}

\begin{proof}
By the triangle inequality and nonexpansiveness,
\[
\begin{aligned}
    \norm{W_{t+1}-V_{t+1}}_{\F}
    &=
    \norm{W_t-\alpha\nabla f(W_t)-V_t+\alpha\nabla f(V_t)+r_t}_{\F}
    \\
    &\leq
    \norm{W_t-\alpha\nabla f(W_t)-V_t+\alpha\nabla f(V_t)}_{\F}
    +\norm{r_t}_{\F}
    \\
    &\leq
    \norm{W_t-V_t}_{\F}+\norm{r_t}_{\F}.
\end{aligned}
\]
Since $V_T=W_T$, iteration gives, for every $t\geq T$,
\[
    \norm{W_t-V_t}_{\F}
    \leq
    \sum_{s=T}^{t-1}\norm{r_s}_{\F}
    \leq
    \sum_{s=T}^{\infty}\norm{r_s}_{\F}
    <\infty,
\]
which proves \cref{eq:bounded-tracking}.

If the stated expansion holds for $V_t$, this uniform bound implies
\[
    W_t
    =V_t+O(1)
    =\widehat W\log t+O(\log\log t).
\]
Since $W_t/\log t$ and $V_t/\log t$ both converge to
$\widehat W\ne0$, we obtain
\[
    \lim_{t\to\infty}\frac{W_t}{\norm{W_t}_{\F}}
    =
    \lim_{t\to\infty}\frac{V_t}{\norm{V_t}_{\F}}
    =
    \frac{\widehat W}{\norm{\widehat W}_{\F}}.
\]
\end{proof}

\subsection{Proof of Theorem~\ref{thm:frobenius}(i)}
\label{app:logistic}

For logistic loss, let $H=R^2/4$ be the global Hessian bound from \cref{lem:logistic-gradient}.
We first establish loss convergence and then compare the iterates with gradient descent to identify the limiting direction.

\begin{proof}
Follow the previous notations: $G_t=\nabla\cL(W_t)$, $g_t=\norm{G_t}_{\F}$, and $P_t=\mathcal P_K\!\left(\frac{G_t}{g_t+\eps}\right)$.

By \cref{lem:logistic-gradient}, $g_t\leq R$.
The stepsize condition $\eta Hu_K\leq\eps$ gives
\[
    1-\frac{H\eta u_K}{2(g_t+\eps)}
    \geq 1-\frac{\eps}{2(g_t+\eps)}
    \geq \frac12.
\]
Since $W_{t+1}=W_t-\eta P_t$, smoothness and
\cref{lem:sector} yield
\begin{align}
    \cL(W_{t+1})
    &\leq \cL(W_t)-\eta\langle G_t,P_t\rangle
    +\frac{H\eta^2}{2}\norm{P_t}_{\F}^2 \notag\\
    &\leq \cL(W_t)
    -\eta\left(1-\frac{H\eta u_K}{2(g_t+\eps)}\right)
    \langle G_t,P_t\rangle \notag\\
    &\leq \cL(W_t)-\frac{\eta}{2}\langle G_t,P_t\rangle
    \notag\\
    &\leq \cL(W_t)-\frac{\eta l_K}{2(g_t+\eps)}g_t^2
    \notag\\
    &\leq \cL(W_t)-\frac{\eta l_K}{2(R+\eps)}g_t^2.
    \label{eq:logistic-descent-proof}
\end{align}

Summing and using $\cL\geq0$ gives
\[
    \sum_{t\geq0}g_t^2
    \leq
    \frac{2(R+\eps)\cL(W_0)}{\eta l_K}
    <\infty.
\]
Hence $g_t\to0$, and \cref{lem:logistic-gradient} implies $\cL(W_t)\to0$.

Set $\alpha=\eta a_K/\eps$ and
\[
    r_t=-\eta\left(P_t-\frac{a_K}{\eps}G_t\right).
\]
Then $W_{t+1}=W_t-\alpha G_t+r_t$.
By \cref{lem:taylor} and the square-summability above,
\[
    \norm{r_t}_{\F}=O(g_t^2),
    \qquad
    \sum_{t\geq0}\norm{r_t}_{\F}<\infty.
\]

Consider the reference iteration
\[
    V_0=W_0,
    \qquad
    V_{t+1}=V_t-\alpha\nabla\cL(V_t).
\]
Since $a_K\leq u_K$, the stepsize condition gives $\alpha\leq1/H$.
The gradient map is therefore globally nonexpansive by \cref{lem:nonexpansive}.
Applying \cref{lem:tracking}, we obtain
\[
    \sup_{t\geq0}\norm{W_t-V_t}_{\F}<\infty.
\]

After vectorization, the reference iteration is gradient descent on the summed logistic loss with stepsize $\alpha/n$.
The summed loss is $nH$-smooth, and $\alpha/n\leq1/(nH)<2/(nH)$.
Thus \citet[Theorem~3]{soudry2018implicit} yields
\[
    V_t=\widehat W_F\log t+O(\log\log t).
\]
By \cref{lem:tracking}, the same expansion holds for $W_t$, and
\[
    \lim_{t\to\infty}\frac{W_t}{\norm{W_t}_{\F}}
    =
    \frac{\widehat W_F}{\norm{\widehat W_F}_{\F}}.
\]
\end{proof}

\subsection{Proof of Theorem~\ref{thm:frobenius}(ii)}
\label{app:exponential}

For exponential loss, we first prove $\cL(W_t)=O(t^{-1})$ and then compare the iterates with gradient descent on a small-loss sublevel set.

\begin{proof} 
Write $\cL_t=\cL(W_t)$. For $s\in[0,1]$, \cref{eq:exp-gradient-bounds},
the displacement bound, and \cref{lem:sector} give
\[
    \norm{\nabla^2\cL(W_t-s\eta P_t)}_{\mathrm{op}}
    \leq R^2\cL_t e^{s\eta R\norm{P_t}_{\F}}
    \leq R^2\cL_t e^{\eta Ru_K}.
\]
Taylor's formula and \cref{lem:sector} therefore yield
\begin{align}
    \cL_{t+1}
    &\leq \cL_t-\eta\langle G_t,P_t\rangle
    +\frac{\eta^2R^2\cL_t e^{\eta Ru_K}}{2}
    \norm{P_t}_{\F}^2 \notag\\
    &\leq \cL_t
    -\eta\left(
        1-\frac{\eta R^2u_K\cL_t e^{\eta Ru_K}}
        {2(g_t+\eps)}
    \right)\langle G_t,P_t\rangle.
    \label{eq:exp-descent-proof}
\end{align}
The stepsize condition and $g_t\geq\gamma\cL_t$ imply
\[
    1-\frac{\eta R^2u_K\cL_t e^{\eta Ru_K}}
    {2(g_t+\eps)}
    \geq
    1-\frac{\gamma\cL_t}{2(g_t+\eps)}
    \geq \frac12.
\]
Thus $\cL_{t+1}\leq\cL_t\leq\cL_0$.
Using the sector lower bound and
$\gamma\cL_t\leq g_t\leq R\cL_t$, we obtain
\begin{align}
    \cL_{t+1}
    &\leq \cL_t-\frac{\eta l_K g_t^2}{2(g_t+\eps)}
    \notag\\
    &\leq \cL_t
    -\frac{\eta l_K\gamma^2\cL_t^2}
    {2(R\cL_t+\eps)} \notag\\
    &\leq \cL_t-c\cL_t^2,
    \qquad
    c=\frac{\eta l_K\gamma^2}{2(R\cL_0+\eps)}.
    \label{eq:exp-loss-recurrence}
\end{align}

Since $\cL_{t+1}>0$, the recurrence implies
$0<c\cL_t<1$, and hence
\[
    \frac{1}{\cL_{t+1}}
    \geq \frac{1}{\cL_t(1-c\cL_t)}
    \geq \frac{1}{\cL_t}+c.
\]
Iteration gives
\[
    \cL_t\leq\frac{1}{\cL_0^{-1}+ct}\longrightarrow0.
\]
Since $e^{-m_i(W_t)}\leq n\cL_t$, every margin tends to infinity.
Moreover, \cref{eq:exp-gradient-bounds} gives
\[
    \sum_{t\geq0}g_t^2
    \leq R^2\sum_{t\geq0}
    \frac{1}{(\cL_0^{-1}+ct)^2}
    <\infty.
\]

Set $\alpha=\eta a_K/\eps$ and write
\[
    W_{t+1}=W_t-\alpha G_t+r_t,
    \qquad
    r_t=-\eta\left(P_t-\frac{a_K}{\eps}G_t\right).
\]
By \cref{lem:taylor} and the square-summability above,
\[
    \norm{r_t}_{\F}=O(g_t^2),
    \qquad
    \sum_{t\geq0}\norm{r_t}_{\F}<\infty.
\]

Choose a finite $T$ such that
$\alpha nR^2\cL_T\leq1/2$, and consider
\[
    V_T=W_T,
    \qquad
    V_{t+1}=V_t-\alpha\nabla\cL(V_t),
    \quad t\geq T.
\]
By \cref{lem:exp-tail-gd} and loss monotonicity,
both trajectories remain in the convex sublevel set
$\{W:\cL(W)\leq\cL_T\}$ for $t\geq T$.
On this set, \cref{eq:exp-gradient-bounds} gives
$\norm{\nabla^2\cL(W)}_{\mathrm{op}}\leq R^2\cL_T$.
Since $\alpha R^2\cL_T\leq1/2<2$,
the gradient map is nonexpansive by \cref{lem:nonexpansive}.

Applying \cref{lem:tracking,lem:exp-tail-gd}, we obtain
\[
    W_t=V_t+O(1)
    =\widehat W_F\log t+O(\log\log t).
\]
Consequently,
\[
    \lim_{t\to\infty}\frac{W_t}{\norm{W_t}_{\F}}
    =
    \frac{\widehat W_F}{\norm{\widehat W_F}_{\F}}.
\]
\end{proof}

\subsection{Proof of Corollary~\ref{cor:momentum-general}}
\label{app:general-momentum}

We first prove summability of the momentum error and then apply the gradient-descent comparison from \cref{app:logistic}.
\begin{proof}
As in \cref{app:logistic}, let $H=R^2/4$. Define the momentum error $e_t=M_t-G_t$ and
\[
    P_t=\mathcal P_K\!\left(
        \frac{N_t}{\norm{N_t}_{\F}+\eps}
    \right).
\]
Then $N_t=G_t+\kappa e_t$ and $W_{t+1}=W_t-\eta P_t$.

By \cref{lem:logistic-gradient}, $\norm{G_t}_{\F}\leq R$.
Since $M_{-1}=0$, the momentum recurrence also gives
$\norm{M_t}_{\F}\leq R$ and $\norm{N_t}_{\F}\leq R$.
Set
\begin{equation}
    c=\frac{l_K}{R+\eps},
    \qquad
    C=\frac{u_K}{\eps},
    \qquad
    \alpha=\frac{\eta a_K}{\eps}.
    \label{eq:momentum-general-constants}
\end{equation}
The stepsize condition \cref{eq:momentum-step} becomes
\[
    \eta HC^2
    \left(\frac{\kappa\beta}{1-\beta}+\frac12\right)
    \leq\frac c2.
\]
By \cref{lem:sector},
\[
    \langle N_t,P_t\rangle
    \geq c\norm{N_t}_{\F}^2,
    \qquad
    \norm{P_t}_{\F}\leq C\norm{N_t}_{\F}.
\]
Global smoothness therefore gives
$\norm{G_{t+1}-G_t}_{\F}\leq\eta HC\norm{N_t}_{\F}$.

The momentum error satisfies
\begin{equation}
    e_t=\beta e_{t-1}+\beta(G_{t-1}-G_t),
    \quad t\geq1,
    \qquad
    e_0=-\beta G_0.
    \label{eq:momentum-lag}
\end{equation}
For a finite horizon $T$, set
\[
S_T=\left(\sum_{t=0}^T\norm{N_t}_{\F}^2\right)^{1/2},
    \qquad
    E_0=\frac{\norm{e_0}_{\F}}{\sqrt{1-\beta^2}}.
\]
Unrolling \cref{eq:momentum-lag} gives
\[
    e_t=\beta^t e_0
    +\sum_{j=0}^{t-1}\beta^{t-j}(G_j-G_{j+1}).
\]
The triangle inequality in the finite-sequence $\ell_2$ norm yields
\begin{align}
    \left(\sum_{t=0}^T\norm{e_t}_{\F}^2\right)^{1/2}
    &\leq E_0+\eta HC
    \sum_{j=1}^T\beta^j
    \left(\sum_{t=0}^{T-j}\norm{N_t}_{\F}^2\right)^{1/2}
    \notag\\
    &\leq E_0+\frac{\beta\eta HC}{1-\beta}S_T.
    \label{eq:momentum-l2-lag}
\end{align}
Since $G_t=N_t-\kappa e_t$, smoothness gives
\begin{align}
    \cL(W_{t+1})
    &\leq\cL(W_t)
    -\eta\langle N_t-\kappa e_t,P_t\rangle
    +\frac{H\eta^2}{2}\norm{P_t}_{\F}^2
    \notag\\
    &\leq\cL(W_t)
    -\eta\left(c-\frac{\eta HC^2}{2}\right)
    \norm{N_t}_{\F}^2
    +\eta\kappa C\norm{e_t}_{\F}\norm{N_t}_{\F}.
    \label{eq:momentum-descent}
\end{align}
Summing, applying Cauchy--Schwarz and
\cref{eq:momentum-l2-lag}, and using
$\cL(W_{T+1})\geq0$, we obtain
\begin{equation}
    \eta\left[
        c-\eta HC^2
        \left(\frac{\kappa\beta}{1-\beta}+\frac12\right)
    \right]S_T^2
    \leq\cL(W_0)+\eta\kappa CE_0S_T.
    \label{eq:momentum-quadratic}
\end{equation}
The bracketed term on the left-hand side is at least $c/2$.
Using
$\kappa CE_0S_T\leq cS_T^2/4+\kappa^2C^2E_0^2/c,$
we conclude that
\[
    S_T^2
    \leq\frac{4\cL(W_0)}{\eta c}
    +\frac{4\kappa^2C^2E_0^2}{c^2}.
\]
Thus $\sum_{t\geq0}\norm{N_t}_{\F}^2<\infty$.
By \cref{eq:momentum-l2-lag} and $G_t=N_t-\kappa e_t$,
\[
    \sum_{t\geq0}\norm{e_t}_{\F}^2<\infty,
    \qquad
    \sum_{t\geq0}\norm{G_t}_{\F}^2<\infty.
\]
Hence $\norm{G_t}_{\F}\to0$.
By \cref{lem:logistic-gradient}, all margins tend to infinity
and $\cL(W_t)\to0$.

We next prove that $\sum_t\norm{e_t}_{\F}<\infty$.
Recall $r_i(W)=(1+e^{\langle A_i,W\rangle})^{-1}$.
Since
$\left|\frac{d}{dz}\log((1+e^z)^{-1})\right|\leq1$,
for any displacement $D$ and $s\in[0,1]$,
\[
    r_i(W+sD)\leq e^{R\norm{D}_{\F}}r_i(W).
\]
Using $r_i(1-r_i)\leq r_i$ and separability, we obtain
\begin{align}
    \norm{\nabla^2\cL(W+sD)}_{\mathrm{op}}
    &\leq\frac{R^2}{n}\sum_i r_i(W+sD)
    \notag\\
    &\leq\frac{R^2}{n}e^{R\norm{D}_{\F}}
    \sum_i r_i(W)
    \notag\\
    &\leq\frac{R^2}{\gamma}e^{R\norm{D}_{\F}}
    \norm{\nabla\cL(W)}_{\F}.
    \label{eq:logistic-relative-curvature}
\end{align}
With $D=-\eta P_t$, the fundamental theorem of calculus
and $\norm{P_t}_{\F}\leq u_K$ give
\begin{align}
    \norm{G_{t+1}-G_t}_{\F}
    &\leq\eta\norm{P_t}_{\F}
    \int_0^1
    \norm{\nabla^2\cL(W_t-s\eta P_t)}_{\mathrm{op}}\,ds
    \notag\\
    &\leq
    \frac{\eta CR^2e^{\eta Ru_K}}{\gamma}
    \norm{G_t}_{\F}\norm{N_t}_{\F}.
    \label{eq:gradient-total-variation}
\end{align}
The right-hand side is summable by Cauchy--Schwarz,
so $\sum_t\norm{G_{t+1}-G_t}_{\F}<\infty$.
Summing the unrolled error recursion therefore yields
\begin{equation}
    \sum_{t\geq0}\norm{e_t}_{\F}
    \leq\frac{\norm{e_0}_{\F}}{1-\beta}
    +\frac{\beta}{1-\beta}
    \sum_{t\geq0}\norm{G_{t+1}-G_t}_{\F}
    <\infty.
    \label{eq:momentum-l1-lag}
\end{equation}
Write the update as
\begin{align*}
    W_{t+1}
    &=W_t-\alpha G_t+r_t,\\
    r_t
    &=-\alpha\kappa e_t
    -\eta\left(P_t-\frac{a_K}{\eps}N_t\right).
\end{align*}
By \cref{lem:taylor},
\[
    \norm{r_t}_{\F}
    \leq\alpha\kappa\norm{e_t}_{\F}
    +O\!\left(\norm{N_t}_{\F}^2\right).
\]
Thus \cref{eq:momentum-l1-lag} and
$\sum_t\norm{N_t}_{\F}^2<\infty$ imply
$\sum_t\norm{r_t}_{\F}<\infty$.

Finally, the stepsize condition gives $\eta HC^2\leq c$.
Since $a_Kl_K\leq u_K^2$,
\[
    \alpha H
    \leq\frac{\eps a_Kl_K}{(R+\eps)u_K^2}
    \leq\frac{\eps}{R+\eps}
    <1.
\]
The gradient map is therefore globally nonexpansive
by \cref{lem:nonexpansive}.
Consider the reference iteration
\[
    V_0=W_0,
    \qquad
    V_{t+1}=V_t-\alpha\nabla\cL(V_t).
\]
Applying \cref{lem:tracking} and the gradient-descent
asymptotics of \citet[Theorem~3]{soudry2018implicit},
as in \cref{app:logistic}, gives
\[
    W_t=V_t+O(1)
    =\widehat W_F\log t+O(\log\log t).
\]
Consequently,
\[
    \lim_{t\to\infty}\frac{W_t}{\norm{W_t}_{\F}}
    =
    \frac{\widehat W_F}{\norm{\widehat W_F}_{\F}}.
\]

\end{proof}

\section{Proof of Section~\ref{sec:limits}}
\label{app:limits}
In this section, we prove the three parts of \cref{thm:limits} in order. Throughout this section, let $r=\min\{d_1,d_2\}$ and $R_*=\max_i\norm{A_i}_*$.

\subsection{Proof of Theorem~\ref{thm:limits}(i)}\label{app:limits-i}
\begin{proof}
For $A\ne0$, the constraints in
\cref{eq:hardmargin,eq:spectral-hardmargin}
are $\ip A W\geq1$. Cauchy--Schwarz gives
\begin{equation*}
\norm A_{\F}\norm W_{\F}\geq\ip A W\geq1.
\end{equation*}
Thus $\norm W_{\F}\geq1/\norm A_{\F}$, with equality
exactly at $\widehat W_{\F}=A/\norm A_{\F}^2$.
By \cref{eq:spectral-margin},
$\gamma_{\F}=\norm A_{\F}$.
For the spectral problem, nuclear/operator duality gives $$\norm A_*\norm W_{\op} \ge \ip A W \ge 1.$$ Since $\norm{\polar(A)}_{\op}=1$ and $\ip A{\polar(A)}=\norm A_*$, \cref{eq:spectral-margin} yields $\gamma_{\op}=\norm A_*$, attained by $\polar(A)$.

We can express the gradient as
$\nabla\cL(W_t)=\ell'(\ip A{W_t})A$ with $\ell'<0$.
At $\eps=0$, we have
\[
\mathcal P_K\left(\frac{\nabla\cL(W_t)}
{\norm{\nabla\cL(W_t)}_{\F}}\right)
=\mathcal P_K\left(-\frac{A}{\norm A_{\F}}\right)
=-\mathcal P_K\left(\frac{A}{\norm A_{\F}}\right)
=-D_K,
\]
where the last step is by the definition of $D_K$ in \cref{thm:limits}(i).
Starting from $W_0=0$, \cref{alg:practical} with $\beta=0$ gives $W_t=(\sum_{s<t}\eta_s)D_K$. Since $\eta_s>0$ for all $s$, normalization gives
$\frac{W_t}{\norm{W_t}_{\F}}=
\frac{D_K}{\norm{D_K}_{\F}}$
for every $t\geq1$, proving \cref{eq:finite-ns-direction}.

We next consider the cubic polynomial. We use any matrix $A$ with at least two distinct positive singular values as a counterexample. Let $\sigma_i>\sigma_j>0$ be two such singular values.

Since $0<\sigma_j<\sigma_i<\norm A_{\F}$ and the cubic polynomial
$p(x)=\frac32x-\frac12x^3$ satisfies
$0<p(x)<1$ and $p'(x)>0$ for $x\in(0,1)$,
we have
\begin{equation}\label{eq:cubic-ratio-contraction}
1<
\frac{p(\sigma_i/\norm A_{\F})}
     {p(\sigma_j/\norm A_{\F})}
=
\frac{\sigma_i}{\sigma_j}
\frac{3-(\sigma_i/\norm A_{\F})^2}
     {3-(\sigma_j/\norm A_{\F})^2}
<
\frac{\sigma_i}{\sigma_j}.
\end{equation}
Thus, for every $K\geq1$,
\[
1<
\frac{p^{\circ K}(\sigma_i/\norm A_{\F})}
     {p^{\circ K}(\sigma_j/\norm A_{\F})}
<
\frac{\sigma_i}{\sigma_j}.
\]
 Thus, the singular values of $D_K$ are unequal and are not proportional to those of $A$. 
Strict Cauchy--Schwarz gives
\[
\frac{\ip{A}{D_K}}{\norm{D_K}_{\F}}
<\norm A_{\F}=\gamma_{\F}.
\]
Moreover,
\[
\frac{\ip{A}{D_K}}{\norm{D_K}_{\op}}
=
\frac{\sum_j \sigma_j p^{\circ K}(\sigma_j/\norm A_{\F})}
{\max_j p^{\circ K}(\sigma_j/\norm A_{\F})}
<\sum_j\sigma_j=\norm A_*=\gamma_{\op}.
\]
Thus, neither margin is achieved. If all positive singular values of $A$ are equal, the Frobenius and compact-polar directions coincide, and every admissible finite polynomial selects that common direction.

Next, we consider the quintic polynomial. Fix any finite $K\geq1$. The invariant interval in \cref{lem:quintic-bounds} implies $p^{\circ K}(1)>0$. Moreover, $p(x)<p'(0)x$ for $0<x\leq5/4$, since
\[
\frac{p(x)}{x}-p'(0)
=x^2(-4.7750+2.0315x^2)<0.
\] Iteration along the invariant interval yields $p^{\circ K}(1)<[p'(0)]^K$.

Take $A=\diag(2,s)$ with $0<s<2$. Then
\[
D_K=\mathcal P_K\left(\frac{A}{\norm A_{\F}}\right)
=\diag\left(
p^{\circ K}\left(\frac{2}{\sqrt{4+s^2}}\right),
p^{\circ K}\left(\frac{s}{\sqrt{4+s^2}}\right)
\right).
\]
Since $(p^{\circ K})'(0)=[p'(0)]^K$,
differentiability at zero and continuity at one give
\begin{equation}\label{eq:quintic-all-k}
\lim_{s\downarrow0}\frac{(D_K)_{11}}{(D_K)_{22}}
=\infty,
\qquad
\lim_{s\downarrow0}\frac{s}{2}
\frac{(D_K)_{11}}{(D_K)_{22}}
=\frac{p^{\circ K}(1)}{[p'(0)]^K}\in(0,1).
\end{equation}
Thus, for sufficiently small $s>0$,
\[
1<\frac{(D_K)_{11}}{(D_K)_{22}}<\frac{2}{s}.
\]
Hence $D_K$ is not proportional to $A$ and has
two unequal positive singular values.
By the same strict margin inequalities as in the
cubic case, $D_K$ achieves neither the Frobenius
nor the spectral max-margin.

For the concrete example $A=\diag(2,1)$,
the Frobenius and spectral hard-margin solutions
are $\widehat W_{\F}=A/5$ and
$\widehat W_{\op}=\Id_2/3$, respectively.
The cubic ratio bound follows directly from
\cref{eq:cubic-ratio-contraction}. For the quintic with $K=5$, we have
\[
D_5=\diag\left(
p^{\circ5}(2/\sqrt5),
p^{\circ5}(1/\sqrt5)
\right).
\]
Direct computation gives
$$
\frac23<(D_5)_{11}<\frac34,\qquad
1<(D_5)_{22}<\frac65.$$
Consequently,
\[
\frac59<\frac{(D_5)_{11}}{(D_5)_{22}}<\frac34.
\]
Since the Frobenius and spectral max-margin directions
have ratios $2$ and $1$, respectively,
$D_5$ achieves neither max-margin.

\end{proof}

\subsection{Proof of Theorem~\ref{thm:limits}(ii)}\label{app:iteration-general}
\begin{proof}
Fix a horizon $T$. Write $S_t=\sum_{s<t}\eta_s$ and $V_t=\sum_{s<t}\eta_s^2$, where $\eta_s=\eta_0/(s+1)$. Then $V_t\leq2\eta_0^2$. Set $G_t^{(K)}=\nabla\cL(W_t^{(K)})$ and $D_t^{(K)}=\mathcal P_K(G_t^{(K)}/(\norm{G_t^{(K)}}_{\F}+\eps))$. The cubic update has operator norm at most one, so $$\norm{W_t^{(K)}}_{\op}\leq\norm{W_0}_{\op}+S_t.$$ Every signed margin up to time $T$ is at most $R_*(\norm{W_0}_{\op}+S_T)$. For either loss, the negative derivative at such a margin is at least $[1+\exp\{R_*(\norm{W_0}_{\op}+S_T)\}]^{-1}$. Taking the inner product of $-G_t^{(K)}$ with the Frobenius-unit separator in \cref{ass:separable} therefore gives
\begin{equation*}
\norm{G_t^{(K)}}_{\F}\geq
\frac{\gamma}{1+\exp\{R_*(\norm{W_0}_{\op}+S_T)\}}>0.
\end{equation*}
Define $\delta_t^{(K)}=1-\ip{G_t^{(K)}}{D_t^{(K)}}/\norm{G_t^{(K)}}_*$ and $E_T^{(K)}=\sum_{t<T}\eta_t\delta_t^{(K)}$. By \cref{lem:dual-oracle-recovery}, uniformly over $t<T$,
\begin{equation}\label{eq:iteration-oracle-bound}
0\leq\delta_t^{(K)}\leq\min\left\{1,\,
 r\left(1+\frac{\eps}{\gamma}
 \left[1+e^{R_*(\norm{W_0}_{\op}+S_T)}\right]\right)(2/3)^K\right\}.
\end{equation}
Consequently, $\lim_{K\to\infty}E_T^{(K)}=0$ for each fixed $T$. 

For the remainder of the proof, we omit the superscript $K$.
Write $Q_t=n^{-1}\sum_i[-\ell'(\ip{A_i}{W_t})]$. 
Both losses satisfy $0<-\ell'\leq\ell$ and $\ell''\leq\ell$.
For any $\norm D_{\op}\leq1$, we have
$|\ip{A_i}D|\leq R_*$, and hence
\[
\frac{d^2}{du^2}\log\cL(W+uD)
\leq
\frac{\frac{d^2}{du^2}\cL(W+uD)}
{\cL(W+uD)}
\leq R_*^2.
\]
Taylor's formula gives
\begin{equation}\label{eq:both-loss-log-descent}
\log \cL(W_{t+1})-\log \cL(W_t)
\leq-\eta_t\frac{\ip{G_t}{D_t}}{\cL(W_t)}
+\frac{R_*^2}{2}\eta_t^2.
\end{equation}
Since $\ip{G_t}{D_t}\geq0$, we have $\cL(W_t)\leq B:=\cL(W_0)e^{R_*^2\eta_0^2}$ at every time. An operator-norm-unit separator attaining $\gamma_{\op}$ in \cref{eq:spectral-margin} gives $\norm{G_t}_*\geq\gamma_{\op}Q_t$.

For exponential loss, $Q_t=\cL(W_t)$. For logistic loss, writing $z=\ell(m)$ gives $-\ell'(m)=1-e^{-z}$, and elementary scalar bounds yield
\begin{equation*}
\frac{z}{1+z}\leq1-e^{-z}\leq z,
\qquad 0\leq z-(1-e^{-z})\leq\frac{z^2}{2}.
\end{equation*}
Each individual loss is at most $n\cL(W_t)$. Averaging these inequalities gives, for both losses,
\begin{equation}\label{eq:logistic-tail-comparison}
\frac{1}{1+nB}\leq\frac{Q_t}{\cL(W_t)}\leq1,
\qquad 0\leq1-\frac{Q_t}{\cL(W_t)}\leq\frac n2\cL(W_t).
\end{equation}
Let $c=\gamma_{\op}/(1+nB)>0$. The alignment bound and \cref{eq:both-loss-log-descent} imply
$$\log \cL(W_t)\leq\log \cL(W_0)-c(S_t-E_t)+R_*^2V_t/2.$$
When $E_T\leq1$, all $t\leq T$ therefore satisfy $$\cL(W_t)\leq Be^c e^{-cS_t}.$$ Because $\eta_t\leq\eta_0$,
\begin{equation*}
\sum_{t<T}\eta_t e^{-cS_t}
\leq e^{c\eta_0}\int_0^{S_T}e^{-cu}\,du
\leq\frac{e^{c\eta_0}}c.
\end{equation*}
It follows that $$\sum_{t<T}\eta_t(1-Q_t/\cL(W_t))\leq nBe^{c(1+\eta_0)}/(2c),$$ uniformly over such $T$ and $K$.
Since
$$
\frac{(1-\delta_t)Q_t}{\cL(W_t)}
\geq 1-\delta_t-
\left(1-\frac{Q_t}{\cL(W_t)}\right),
$$
applying this bound to \cref{eq:both-loss-log-descent}
yields
\begin{equation*}
-\log \cL(W_T)\geq\gamma_{\op}S_T-\gamma_{\op}E_T-C_0,
\end{equation*}
where $C_0$ is a finite constant independent of $T$ and $K$.
For exponential loss, $$\min_i\ip{A_i}{W_T}\geq-\log(n\cL(W_T)).$$ For logistic loss, each individual loss is at most $n\cL(W_T)$, so
\begin{equation*}
\min_i\ip{A_i}{W_T}\geq-\log(e^{n\cL(W_T)}-1)
\geq-\log(n\cL(W_T))-n\cL(W_T).
\end{equation*}
The last inequality follows from $e^x-1\leq xe^x$ for $x>0$. Since $\cL(W_T)\leq B$, either loss gives
$$\min_i\ip{A_i}{W_T}\geq\gamma_{\op}S_T-\gamma_{\op}E_T-C$$
where  $C$ is a finite positive constant independent of $T,K$ whenever $E_T\leq1$.

For all sufficiently large $T$ and then all sufficiently large $K$, this lower bound is positive. Using \cref{eq:spectral-margin}, dividing by $\norm{W_T^{(K)}}_{\op}\leq\norm{W_0}_{\op}+S_T$, and taking $K\to\infty$ at this fixed $T$ gives
\begin{equation}\label{eq:iteration-finite-horizon}
0\leq\limsup_{K\to\infty}
\left(\gamma_{\op}-\frac{\min_i\ip{A_i}{W_T^{(K)}}}{\norm{W_T^{(K)}}_{\op}}\right)
\leq\frac{\gamma_{\op}\norm{W_0}_{\op}+C}{\norm{W_0}_{\op}+S_T}.
\end{equation}
Finally, $S_T\geq\eta_0\log(T+1)$, so the right-hand side tends to zero as $T\to\infty$. Setting $\eps=0$ proves the cubic assertion in part (ii) for both losses and every finite $W_0$.

For the quintic, take $n=1$, $A=\diag(2,1)$, $W_0=0$, and $\eps=0$. By \cref{lem:quintic-cycle}, the diagonal-entry ratios of the matrices $D_K$ from \cref{thm:limits}(i) lie in two disjoint intervals for odd and even $K\geq3$. The update identity in \cref{app:limits-i} gives, for every fixed $t\geq1$ and any positive stepsize sequence,
\[
\frac{W_t^{(K)}}{\norm{W_t^{(K)}}_{\F}}
=\frac{D_K}{\norm{D_K}_{\F}}.
\]
Thus the normalized iterates do not converge as $K\to\infty$.
\end{proof}

\subsection{Proof of Theorem~\ref{thm:limits}(iii)}\label{app:limits-iii}
\begin{proof}
Fix $\eps>0$. The proof in \cref{app:iteration-general} treats every fixed $\eps\geq0$. In particular, the factor multiplying $(2/3)^K$ in \cref{eq:iteration-oracle-bound} is finite at each fixed horizon $T$, so $\lim_{K\to\infty}E_T^{(K)}=0$. Thus \cref{eq:iteration-finite-horizon} holds for the present stabilizer. Since $S_T\geq\eta_0\log(T+1)$, taking $T\to\infty$ gives \cref{eq:iteration-general}.
\end{proof}

\subsection{Auxiliary lemmas}\label{app:limits-auxiliary}

\begin{lemma}\label{lem:cubic-recovery}
For every $s\in(0,1]$ and integer $K\geq0$, the cubic in \cref{equ:cubic} satisfies
\begin{equation}
 p_{\mathrm{cub}}^{\circ K}(s)\geq\frac{s}{\sqrt{s^2+(2/3)^{2K}(1-s^2)}},
 \qquad 0\leq s(1-p_{\mathrm{cub}}^{\circ K}(s))\leq(2/3)^K.
 \label{eq:cubic-weighted-error}
\end{equation}
Moreover, $0\leq p_{\mathrm{cub}}^{\circ K}(s)\leq1$ on $[0,1]$.
\end{lemma}
\begin{proof}
We omit the subscript $\mathrm{cub}$ in this proof. The cubic polynomial $p$ maps \([0,1]\) into \([0,1]\) and is strictly positive on \((0,1]\). For $0<s<1$,
\begin{equation}
\begin{aligned}
\frac{p(s)^{-2}-1}{s^{-2}-1}
&=\frac{(1-s^2)(4-s^2)}{(3-s^2)^2}\\
&=\frac49-\frac{s^2(21-5s^2)}{9(3-s^2)^2}
\leq\frac49.
\end{aligned}
\label{eq:cubic-error-contraction}
\end{equation}
At $s=1$, $p(s)=1$. Iterating the bound yields
$p^{\circ K}(s)^{-2}-1\leq(4/9)^K(s^{-2}-1)$, which is the first bound in \cref{eq:cubic-weighted-error}.

Since $1-s^2\leq1$, that bound also gives $p^{\circ K}(s)\geq s/\sqrt{s^2+(2/3)^{2K}}$. Consequently

\begin{equation*}
s(1-p^{\circ K}(s))\leq s(1-s/\sqrt{s^2+(2/3)^{2K}})\leq\sqrt{s^2+(2/3)^{2K}}-s\leq(2/3)^K,
\end{equation*}
where the second inequality uses $s\leq\sqrt{s^2+(2/3)^{2K}}$, and the last follows from $\sqrt{s^2+(2/3)^{2K}}\leq s+(2/3)^K$. 
\end{proof}

\begin{lemma}[Relative dual-oracle error]\label{lem:dual-oracle-recovery}
Let $G\ne0$, $\varepsilon\geq0$, and
$D=\mathcal P_K(G/(\norm G_{\F}+\varepsilon))$ with the cubic polynomial. Then $\|D\|_{\mathrm{op}}\leq1$ and
\begin{equation}
 0\leq\|G\|_*-\langle G,D\rangle
 \leq r(\norm G_{\F}+\varepsilon)(2/3)^K,
 \qquad
 \langle G,D\rangle\geq
 \left[1-r\left(1+\frac\varepsilon{\norm G_{\F}}\right)(2/3)^K\right]\|G\|_*.
 \label{eq:dual-oracle-recovery}
\end{equation}
\end{lemma}
\begin{proof}
Let $G=U\Sigma V^\top$ be a compact SVD and set $s_j=\sigma_j(G)/(\norm G_{\F}+\varepsilon)\in(0,1]$. The singular values of $D$ are $p_{\mathrm{cub}}^{\circ K}(s_j)\in(0,1]$, so $\norm D_{\op}\leq1$. By \cref{lem:cubic-recovery},
\begin{align*}
0\leq\norm G_*-\ip G D
&=\sum_j\sigma_j(G)(1-p_{\mathrm{cub}}^{\circ K}(s_j))\\
&=(\norm G_{\F}+\varepsilon)\sum_j s_j(1-p_{\mathrm{cub}}^{\circ K}(s_j))\\
&\leq r(\norm G_{\F}+\varepsilon)(2/3)^K.
\end{align*}
Dividing by $\norm G_*\geq\norm G_{\F}$ gives the relative bound.
\end{proof}

\begin{lemma}[Bounds for the quintic]\label{lem:quintic-bounds}
The quintic in \cref{eq:practical-polynomial} satisfies
\begin{equation}\label{eq:quintic-trapping}
\begin{aligned}
p_{\mathrm{quin}}([0,5/4])&\subseteq[0,5/4],&
p_{\mathrm{quin}}([1/2,5/4])&\subseteq[1/2,5/4],\\
p_{\mathrm{quin}}(s)&\geq2s\quad(0\leq s\leq1/2),&
p_{\mathrm{quin}}(s)&\leq p_{\mathrm{quin}}'(0)s\quad(0\leq s\leq5/4).
\end{aligned}
\end{equation}
\end{lemma}
\begin{proof}
We omit the subscript $\mathrm{quin}$ in this proof. For a polynomial $h$ of degree $d$ and a rational interval $[l,u]$, write its Bernstein expansion as
\begin{equation}\label{eq:bernstein-certificate}
h(l+(u-l)x)=\sum_{j=0}^d b_j\binom dj x^j(1-x)^{d-j}.
\end{equation}
On $[0,1]$, the nonnegative basis functions sum to one, so
$\min_j b_j\leq h(l+(u-l)x)\leq\max_j b_j$.
Applying this bound to the six coefficients of $p$ on each interval gives
\begin{equation*}
\begin{aligned}
p([0,1/2])&\subseteq[0,6/5],&
p([1/2,3/4])&\subseteq[1,5/4],\\
p([3/4,1])&\subseteq[7/10,11/10],&
p([1,5/4])&\subseteq[3/5,6/5].
\end{aligned}
\end{equation*}
These enclosures prove both interval inclusions in \cref{eq:quintic-trapping}. For $0<s\leq1/2$, $$p(s)/s\geq p'(0)-4.775/4>2.$$ For $0<s\leq5/4$,
$$p(s)/s-p'(0)=s^2(-4.775+2.0315s^2)<0$$ At $s=0$, both scalar bounds hold with equality.
\end{proof}
\begin{lemma}[Quintic iteration for $\diag(2,1)$]\label{lem:quintic-cycle}
Let $A=\diag(2,1)$ and $D_K=\mathcal P_K(A/\norm{A}_{\F})$
for the quintic in \cref{eq:practical-polynomial}.
Then $D_K/\norm{D_K}_{\F}$ does not converge as $K\to\infty$.
\end{lemma}

\begin{proof}
We omit the subscript $\mathrm{quin}$ for $p$ in this proof.
We show that the two diagonal entries of $D_K$ alternate between
two disjoint intervals, so their ratio cannot converge.

Set
\[
I_-=[2/3,4/5],\qquad I_+=[19/20,23/20].
\]
The Bernstein bound in \cref{eq:bernstein-certificate},
applied on $I_-$, $[19/20,1]$, and $[1,23/20]$, gives
\[
p(I_-)\subset I_+,\qquad p(I_+)\subset I_-.
\]

For the two normalized diagonal entries of $A$, we calculate
the first three scalar iterates.
Since $2/\sqrt5\in[8/9,9/10]$ and
$1/\sqrt5\in[4/9,9/20]$, applying the same Bernstein
bound three times gives
\begin{equation}\label{eq:quintic-cycle-entries}
p^{\circ3}([8/9,9/10])\subset I_-,
\qquad
p^{\circ3}([4/9,9/20])\subset I_+.
\end{equation}

Thus the diagonal entries of $D_3$ lie in $I_-$ and $I_+$,
respectively, and every subsequent iteration exchanges
these intervals. Consequently, for every $K\geq3$,
\begin{equation}\label{eq:quintic-matrix-cycle}
\frac{(D_K)_{11}}{(D_K)_{22}}\in
\begin{cases}
[40/69,16/19],&K\text{ odd},\\
[19/16,69/40],&K\text{ even}.
\end{cases}
\end{equation}
Since these two intervals are disjoint, the normalized
matrices cannot converge to a common limit along the
odd and even subsequences.
\end{proof}
\section{Proof of Section~\ref{sec:recovery}}
\label{app:recovery}
We prove the cubic recovery guarantee with fixed stabilization and the quintic recovery guarantee with randomized stabilization. We use $r$ and $R_*$ from \cref{app:limits}.

\subsection{Proof of Theorem~\ref{thm:recovery}}\label{app:recovery-proof}
\begin{proof}
Let $S_T,E_T,V_T$ be as in \cref{lem:inexact-spectral-recovery}. By \cref{lem:dual-oracle-recovery}, the cubic update
$D_t=\mathcal P_{K_t}(G_t/(\norm{G_t}_{\F}+\eps))$
satisfies $\norm{D_t}_{\op}\leq1$. Thus $\norm{W_t}_{\op}\leq\norm{W_0}_{\op}+S_t$. For $t\geq0$, the harmonic stepsize and $\eta_0R_*\leq1$ give $R_*S_t\leq1+\log(t+1)$. Each margin is at most $R_*(\norm{W_0}_{\op}+S_t)$, so \cref{eq:exp-gradient-bounds} yields
\begin{equation}\label{eq:fixed-eps-gradient-lower}
\norm{G_t}_{\F}\geq\gamma\cL(W_t)
\geq\gamma e^{-R_*(\norm{W_0}_{\op}+S_t)}\geq\frac{\gamma e^{-1-R_*\norm{W_0}_{\op}}}{t+1}.
\end{equation}
Let $\delta_t=1-\ip{G_t}{D_t}/\norm{G_t}_*$. The positive cubic singular values imply $0\leq\delta_t\leq1$. Since $K_t=\left\lceil\frac{2\log(t+2)}{\log(3/2)}\right\rceil$, \cref{lem:dual-oracle-recovery,eq:fixed-eps-gradient-lower} imply
\begin{equation}\label{eq:fixed-eps-oracle-bound}
\delta_t\leq\min\left\{1,\frac{r(1+\eps e^{1+R_*\norm{W_0}_{\op}}(t+1)/\gamma)}{(t+2)^2}\right\}
\leq\frac{r(1+\eps e^{1+R_*\norm{W_0}_{\op}}/\gamma)}{t+2}.
\end{equation}
Consequently, using $\sum_{t\geq0}[(t+1)(t+2)]^{-1}=1$,
\begin{equation}\label{eq:fixed-eps-error-sum}
E_T\leq\eta_0r(1+\eps e^{1+R_*\norm{W_0}_{\op}}/\gamma),\qquad
V_T\leq2\eta_0^2,\qquad S_T\geq\eta_0\log(T+1).
\end{equation}
The bound from
\cref{lem:inexact-spectral-recovery} gives
\[
-\log\cL(W_T)
\geq-\log\cL(W_0)+\gamma_{\op}S_T
-\gamma_{\op}E_T-R_*^2V_T/2.
\]
Since $S_T\to\infty$ while $E_T$ and $V_T$
remain bounded, the loss converges to zero
and the positivity condition in
\cref{lem:inexact-spectral-recovery}
holds for sufficiently large $T$.
 The resulting explicit margin bound is
\begin{equation}\label{eq:fixed-eps-margin-bound}
\begin{aligned}
0&\leq\gamma_{\op}-\frac{\min_i\ip{A_i}{W_T}}{\norm{W_T}_{\op}}\\
&\leq\frac{\gamma_{\op}\eta_0r(1+\eps e^{1+R_*\norm{W_0}_{\op}}/\gamma)+R_*^2\eta_0^2
+\log(n\cL(W_0))+\gamma_{\op}\norm{W_0}_{\op}}
{\eta_0\log(T+1)+\norm{W_0}_{\op}}.
\end{aligned}
\end{equation}
This gives \cref{eq:recovery-margin} as $T\to\infty$.
\end{proof}

\subsection{Proof of Theorem~\ref{thm:quintic-recovery}}\label{app:quintic-recovery}
\begin{proof}
We omit the subscript $\mathrm{quin}$ in this proof. Let $\mathcal F_t$ be the history generated by $U_0,\ldots,U_{t-1}$. The matrix $G_t$ is $\mathcal F_t$-measurable and is nonzero by \cref{eq:exp-gradient-bounds}. Let $\mu_t$ be the scalar mean defined in \cref{lem:quintic-average}. Write
\begin{equation*}
D_t=\mathcal P_{K_t}\!\left(\frac{G_t}{\norm{G_t}_{\F}+\eps_t}\right),\qquad
\overline D_t=\mathbb E[D_t\mid\mathcal F_t].
\end{equation*}
The schedule in~\cref{thm:quintic-recovery} gives $G_t/(\norm{G_t}_{\F}+\eps_t)=(t+2)^{-U_t}G_t/\norm{G_t}_{\F}$. Since $U_t$ is independent of $\mathcal F_t$, \cref{lem:quintic-average,lem:quintic-mean-update} apply conditionally on $\mathcal F_t$. Separability and nuclear/operator duality give
\begin{equation}\label{eq:quintic-loss-alignment}
0<\gamma_{\op}\cL(W_t)\leq\norm{G_t}_*\leq R_*\cL(W_t),\qquad
\frac{\ip{G_t}{\overline D_t}}{\cL(W_t)}
\geq\mu_t\gamma_{\op}-\frac{5rR_*}{4e\log(t+2)},
\end{equation}
where we used $\norm{G_t}_{\F}\leq\norm{G_t}_*\leq R_*\cL(W_t)$ to bound the error term.

Since $\norm{D_t}_{\F}\leq5\sqrt r/4$, the matrix martingale $\sum_{t<T}\eta_t(D_t-\overline D_t)$ has total expected squared increments at most $\frac{25r}{4}\sum_t\eta_t^2<\infty$. It is therefore bounded in $L^2$ and converges almost surely. Decomposing
\begin{equation*}
W_T=W_0-\sum_{t<T}\eta_t\overline D_t-\sum_{t<T}\eta_t(D_t-\overline D_t)
\end{equation*}
and using $\norm{\overline D_t}_{\op}\leq\mu_t+p'(0)/\log(t+2)$ from \cref{lem:quintic-mean-update} gives, for a finite nonnegative random constant $C_1$,
\begin{equation}\label{eq:quintic-predictor-upper}
\norm{W_T}_{\op}\leq
\sum_{t<T}\eta_t\mu_t+p'(0)\sum_{t<T}\frac{\eta_t}{\log(t+2)}+C_1
\quad\text{for every }T\geq1
\end{equation}
almost surely.

The bound used in \cref{eq:recovery-logloss}, now with $\norm{D_t}_{\op}\leq5/4$, gives
\begin{equation}\label{eq:quintic-logloss-step}
-\log\cL(W_{t+1})\geq-\log\cL(W_t)
+\eta_t\frac{\ip{G_t}{D_t}}{\cL(W_t)}
-\frac{25R_*^2}{32}\eta_t^2.
\end{equation}
Its quadratic term follows from the variance bound for a variable in $[-5R_*/4,5R_*/4]$. The scalar martingale
\begin{equation*}
\sum_{t<T}\eta_t\frac{\ip{G_t}{D_t-\overline D_t}}{\cL(W_t)}
\end{equation*}
also converges almost surely, because its increments have absolute values at most $5R_*\eta_t/2$ and $\sum_t\eta_t^2<\infty$. Summing~\cref{eq:quintic-logloss-step} and using~\cref{eq:quintic-loss-alignment} therefore gives a finite nonnegative random constant $C_2$ such that
\begin{equation}\label{eq:quintic-logloss-lower}
-\log\cL(W_T)\geq\gamma_{\op}\sum_{t<T}\eta_t\mu_t
-\frac{5rR_*}{4e}\sum_{t<T}\frac{\eta_t}{\log(t+2)}-C_2
\quad\text{for every }T\geq1
\end{equation}
almost surely, and the constant $C_2$ includes the bounded quadratic sum and $|\log\cL(W_0)|$. Set $$A_T=\sum_{t<T}\eta_t\mu_t, \qquad B_T=\sum_{t<T}\eta_t/\log(t+2).$$ Since $\mu_t\geq1/4$ and $\eta_t=\eta_0/(t+1)$,
\begin{equation}\label{eq:quintic-averaging-sums}
A_T\geq\frac{\eta_0}{4}\log(T+1),\qquad
B_T=O(1+\log\log(T+2))\quad\text{as }T\to\infty.
\end{equation}
Thus the right-hand side of~\cref{eq:quintic-logloss-lower} diverges to infinity, establishing loss convergence. Every exponential term is at most $n\cL(W_T)$, so the minimum margin is at least $-\log\cL(W_T)-\log n$ and is eventually positive. Combining \cref{eq:quintic-predictor-upper,eq:quintic-logloss-lower} then yields, almost surely for all sufficiently large $T$,
\begin{equation}\label{eq:quintic-explicit-gap}
0\leq\gamma_{\op}-\frac{\min_i\ip{A_i}{W_T}}{\norm{W_T}_{\op}}
\leq\frac{(p'(0)\gamma_{\op}+5rR_*/(4e))B_T+\gamma_{\op}C_1+C_2+\log n}
{A_T+p'(0)B_T+C_1},
\end{equation}
where the lower bound follows from the definition of $\gamma_{\op}$. Substitution of~\cref{eq:quintic-averaging-sums} proves~\cref{eq:quintic-random-margin}.
\end{proof}

\subsection{Auxiliary lemmas}\label{app:recovery-auxiliary}

\begin{lemma}[Inexact spectral margin]
\label{lem:inexact-spectral-recovery}
Suppose \cref{ass:separable} holds and consider
the exponential loss.
Let $W_{t+1}=W_t-\eta_tD_t$ with finite $W_0$,
where $\eta_t>0$, $\norm{D_t}_{\op}\leq1$, and
$\ip{G_t}{D_t}\geq(1-\delta_t)\norm{G_t}_*$
for some $0\leq\delta_t\leq1$.
Set $S_T=\sum_{t<T}\eta_t$,
$E_T=\sum_{t<T}\eta_t\delta_t$,
and $V_T=\sum_{t<T}\eta_t^2$.
Whenever $$\gamma_{\op}(S_T-E_T)-R_*^2V_T/2
-\log(n\cL(W_0))>0,$$ we have
\begin{equation}
\label{eq:inexact-spectral-margin}
0\leq\gamma_{\op}
-\frac{\min_i\ip{A_i}{W_T}}{\norm{W_T}_{\op}}
\leq
\frac{
\gamma_{\op}E_T+R_*^2V_T/2
+\log(n\cL(W_0))
+\gamma_{\op}\norm{W_0}_{\op}
}{
S_T+\norm{W_0}_{\op}
}.
\end{equation}
\end{lemma}

\begin{proof}
For each $t$, define a discrete random variable $X_t$
that takes value $\ip{A_i}{D_t}$ with probability $
\frac{e^{-\ip{A_i}{W_t}}}{n\cL(W_t)},
\quad i\in[n].$  These weights sum to one, and nuclear/operator duality gives $|X_t|\leq R_*$. Also
$\mathbb E X_t=-\langle G_t,D_t\rangle/\mathcal L(W_t)$.
The exponential-loss identity and the bounded-variable moment bound give
\begin{align}
 \log\mathcal L(W_{t+1})-\log\mathcal L(W_t)
 &=\log\mathbb E e^{\eta_tX_t}\leq-\eta_t\frac{\langle G_t,D_t\rangle}{\mathcal L(W_t)}
       +\frac{\eta_t^2R_*^2}{2}.
 \label{eq:recovery-logloss}
\end{align}
To obtain the moment bound, let $F(a)=\log\mathbb E e^{aX}$ for $X\in[-R_*,R_*]$. The derivative $F''(a)$ is the variance under the exponentially tilted law and is bounded by its second moment, hence by $R_*^2$. Taylor's formula at zero gives $F(a)\leq a\mathbb EX+a^2R_*^2/2$ for $a\geq0$.

Let $W_{\op}$ attain $\gamma_{\op}$ in
\cref{eq:spectral-margin}. By duality,
\[
\begin{aligned}
\norm{G_t}_*
&\geq \ip{-G_t}{W_{\op}}\\
&=\frac1n\sum_i e^{-\ip{A_i}{W_t}}
\ip{A_i}{W_{\op}}\\
&\geq\gamma_{\op}\frac1n
\sum_i e^{-\ip{A_i}{W_t}}\\
&=\gamma_{\op}\cL(W_t).
\end{aligned}
\]
Using $1-\delta_t\geq0$ in \cref{eq:recovery-logloss} and summing gives
\begin{equation}
-\log\mathcal L(W_T)\geq-\log\mathcal L(W_0)+\gamma_{\mathrm{op}}S_T-\gamma_{\mathrm{op}}E_T-\frac{R_*^2}{2}V_T.
\end{equation}

Each exponential term is at most $n\mathcal L(W_T)$, so the minimum margin is at least $-\log\mathcal L(W_T)-\log n$. Also $\|W_T\|_{\mathrm{op}}\leq\norm{W_0}_{\op}+S_T$ by the triangle inequality and the oracle norm bound. The stated positivity condition ensures that the minimum margin is positive and $W_T\ne0$. Dividing this positive lower bound by $\|W_T\|_{\mathrm{op}}\leq\norm{W_0}_{\op}+S_T$ proves the right-hand inequality in \cref{eq:inexact-spectral-margin}. The left-hand inequality is the definition of the spectral max-margin, applied to $W_T/\|W_T\|_{\mathrm{op}}$.
\end{proof}

\begin{lemma}[Logarithmic averaging of the quintic]\label{lem:quintic-average}
For each integer $t\geq0$, let $K_t$ be the iteration count in \cref{thm:quintic-recovery}, let $U_t\sim\operatorname{Unif}(0,1)$, and define
$\mu_t=\int_0^1p_{\mathrm{quin}}^{\circ K_t}((t+2)^{-u})\,du$.
Then $1/4\leq\mu_t\leq5/4$, and for every $0<s\leq1$,
\begin{equation}\label{eq:quintic-scalar-average}
\mu_t-\frac{5\log(1/s)}{4\log(t+2)}
\leq\mathbb E p_{\mathrm{quin}}^{\circ K_t}(s(t+2)^{-U_t})
\leq\mu_t+\frac{p_{\mathrm{quin}}'(0)}{\log(t+2)}.
\end{equation}
\end{lemma}
\begin{proof}
We omit the subscript $\mathrm{quin}$ in this proof. By \cref{lem:quintic-bounds}, an orbit starting at $x\in[(t+2)^{-1/2},1]$ and staying below $1/2$ through iteration $K_t$ would satisfy
$p^{\circ K_t}(x)\geq2^{K_t}(t+2)^{-1/2}>1$, since $p'(0)=3.4445<4$.
Hence it enters $[1/2,5/4]$ by iteration $K_t$ and stays there. Consequently,
\[
\frac14\leq\int_0^{1/2}p^{\circ K_t}((t+2)^{-u})\,du
\leq\mu_t\leq\frac54.
\]
A change of variables gives
\begin{equation}\label{eq:quintic-average-shift}
\begin{aligned}
&\mathbb E p^{\circ K_t}(s(t+2)^{-U_t})-\mu_t\\
&\quad=\frac{1}{\log(t+2)}\left(
\int_{\log(t+2)}^{\log(t+2)+\log(1/s)}p^{\circ K_t}(e^{-u})\,du
-\int_0^{\log(1/s)}p^{\circ K_t}(e^{-u})\,du\right).
\end{aligned}
\end{equation}
The bounds $0\leq p^{\circ K_t}\leq5/4$ on $[0,1]$ from \cref{lem:quintic-bounds} give the lower inequality in \cref{eq:quintic-scalar-average}. Iterating the linear upper bound in the same lemma gives $p^{\circ K_t}(x)\leq[p'(0)]^{K_t}x$ for $0\leq x\leq1$, so
\[
\begin{aligned}
\int_{\log(t+2)}^{\log(t+2)+\log(1/s)}p^{\circ K_t}(e^{-u})\,du
&\leq\int_{\log(t+2)}^\infty[p'(0)]^{K_t}e^{-u}\,du\\
&=\frac{[p'(0)]^{K_t}}{t+2}\leq p'(0).
\end{aligned}
\]
Substitution into \cref{eq:quintic-average-shift} proves the upper inequality.
\end{proof}

\begin{lemma}[Mean quintic update]\label{lem:quintic-mean-update}
For each integer $t\geq0$, use the quintic and iteration count $K_t$ from \cref{thm:quintic-recovery}. For $G\ne0$, let $U_t\sim\operatorname{Unif}(0,1)$ and
$D=\mathcal P_{K_t}((t+2)^{-U_t}G/\norm G_{\F})$.
Then $\norm D_{\op}\leq5/4$ and, with $\mu_t$ defined in \cref{lem:quintic-average},
\begin{equation}\label{eq:quintic-mean-oracle}
\norm{\mathbb E D}_{\op}\leq\mu_t+\frac{p_{\mathrm{quin}}'(0)}{\log(t+2)},\qquad
\ip G{\mathbb E D}\geq\mu_t\norm G_*-\frac{5r}{4e\log(t+2)}\norm G_{\F}.
\end{equation}
\end{lemma}
\begin{proof}
Let $G=U\Sigma Q^\top$ be a compact SVD, with $s_j=\sigma_j(G)/\norm G_{\F}\in(0,1]$. The transformation preserves these singular vectors, so \cref{lem:quintic-average} bounds each singular value of its expectation. The invariant interval in \cref{lem:quintic-bounds} gives $\norm D_{\op}\leq5/4$. Summing the lower scalar bounds with weights $\sigma_j(G)$ gives
\[
\begin{aligned}
\ip G{\mathbb E D}
&\geq\mu_t\norm G_*-\frac{5\norm G_{\F}}{4\log(t+2)}\sum_j s_j\log(1/s_j)\\
&\geq\mu_t\norm G_*-\frac{5r}{4e\log(t+2)}\norm G_{\F},
\end{aligned}
\]
where the last step is by $s\log(1/s)\leq1/e$ on $(0,1]$.
\end{proof}

\section{Additional experiments}\label{app:experiments}

\subsection{Synthetic Dataset Construction}
\label{app:matrix-data}

We construct the five synthetic datasets in
\cref{sec:matrix-experiments} around known
Frobenius hard-margin solutions.
For each dataset, we draw independent orthogonal
matrices $Q,V\in\R^{4\times4}$ by QR factorization
of independent Gaussian matrices and define
\[
W_\star=Q\diag(2,1,0.7,0.3)V^\top,
\]
We generate eight independent Gaussian matrices,
subtract their mean, and project each orthogonally
to $W_\star$ in the Frobenius inner product.
We rescale the resulting matrices $H_i$ so that
their mean squared Frobenius norm is one, and set $
A_i=\frac{W_\star}{\norm{W_\star}_{\F}^2}+0.32H_i,\quad i=1,\ldots,8.$
By construction,
$\sum_{i=1}^8H_i=0$ and
$\ip{H_i}{W_\star}=0$.
For the remaining four samples, we set $
A_i=1.6\frac{W_\star}{\norm{W_\star}_{\F}^2}+0.15E_i,\quad i=9,\ldots,12,$
where the $E_i$ are independent Gaussian matrices
projected orthogonally to $W_\star$.
Thus, the first eight samples have margin one
at $W_\star$, while the remaining four have
margin $1.6$.
We independently sample labels
$y_i\sim\operatorname{Unif}\{-1,+1\}$
and set $X_i=y_iA_i$. For any feasible $W$ in \cref{eq:hardmargin}, summing the first eight constraints gives
\begin{equation*}
1\leq\frac18\sum_{i=1}^8\ip{A_i}{W}
=\frac{\ip{W_\star}{W}}{\norm{W_\star}_{\F}^2}
\leq\frac{\norm W_{\F}}{\norm{W_\star}_{\F}}.
\end{equation*}
Since $W_\star$ is feasible, it minimizes the Frobenius norm. Equality in Cauchy--Schwarz makes it the unique minimizer. The spectral hard-margin solutions are computed
using CVXPY with the Clarabel solver.

\subsection{Additional experiment on quintic and momentum}
\label{app:momentum-experiments}
\begin{figure}[t]
\centering
\includegraphics[width=.91\textwidth]{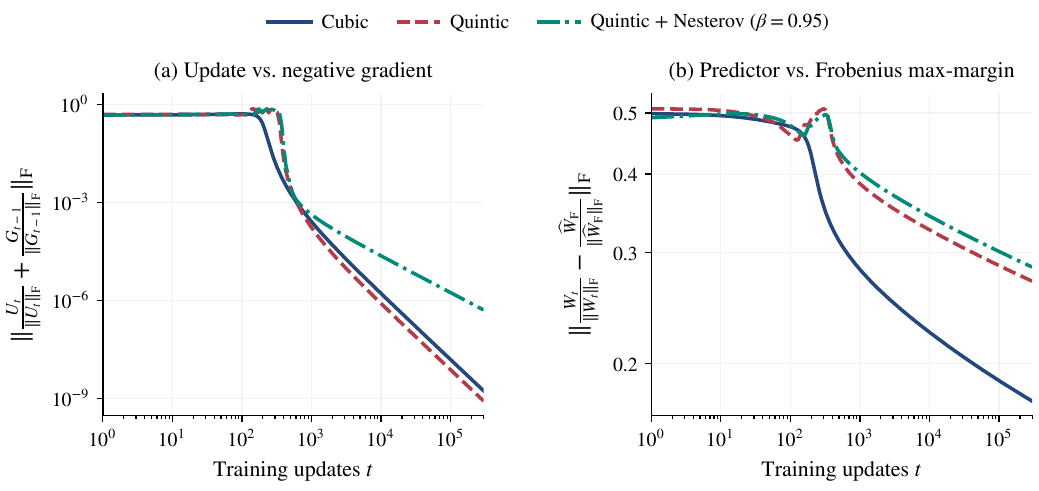}
\caption{
Direction distances during Muon training with
cubic, quintic, and quintic with Nesterov-style
momentum, under the experimental settings
described in \cref{app:momentum-experiments}.
Results are averaged over five random seeds.
}
\label{fig:momentum}
\end{figure}

We compare three variants of \cref{alg:practical}:
cubic, quintic, and quintic with Nesterov-style
momentum (\cref{eq:practical-momentum}).
All methods use $K=5$, fixed $\eps=10^{-3}$,
and constant stepsize $\eta=0.05$ for
$3\times10^5$ updates on the same five datasets
as in \cref{sec:matrix-experiments}.
The momentum variant uses $\beta=\kappa=0.95$,
while the other two use $\beta=0$. We measure the distance between the normalized
parameter update $U_t=W_t-W_{t-1}$ and the
negative gradient $-G_{t-1}$, as well as the
distance between the normalized predictor $W_t$
and the Frobenius max-margin direction
$\widehat W_{\F}/\norm{\widehat W_{\F}}_{\F}$.
Results are averaged over five random seeds. \Cref{fig:momentum}(a) shows that the update
directions approach the negative-gradient direction
much faster than the predictors approach the
Frobenius max-margin direction in panel (b).
This observation is consistent with the local
gradient approximation underlying
\cref{thm:frobenius}.
Momentum affects the transient behavior,
while the two quintic variants exhibit
similar predictor trajectories.

\end{document}